\pdfoutput=1
\documentclass[11pt]{article}

\usepackage[]{ACL2023}

\usepackage{times}
\usepackage{latexsym}

\usepackage[T1]{fontenc}
\usepackage[utf8]{inputenc}

\usepackage{microtype}

\usepackage{inconsolata}

\usepackage{booktabs}
\usepackage{todonotes}
\usepackage{multirow}
\usepackage{amssymb} % for checkmark
\usepackage[inline, shortlabels]{enumitem}
\usepackage{microtype}
\usepackage{xcolor}
\definecolor{YellowOrange}{RGB}{247, 158, 22}
\definecolor{BrickRed}{rgb}{0.8, 0.25, 0.33}
\definecolor{persimmon}{rgb}{0.93, 0.35, 0.0}
\definecolor{darkcyan}{rgb}{0.0, 0.55, 0.55}
\definecolor{darkorange}{rgb}{1.0, 0.55, 0.0}
\usepackage{array}
\newcolumntype{P}[1]{>{\hspace{0pt}}p{#1}}
\usepackage{siunitx}
\newcommand{\cent}[1]{\num[round-mode=places,round-precision=1]{#1}\%}
\newcommand{\float}[1]{\num[round-mode=places,round-precision=2]{#1}}

\renewcommand\paragraph[1]{{\normalsize\bf #1}}

\newcommand{\Yes}{\textsc{Yes}}
\newcommand{\No}{\textsc{No}}
\newcommand{\NA}{\textsc{N/A}}
\newcommand{\Blank}{\textsc{Blank}}

\newcommand{\Findings}{\textsc{Findings}}
\newcommand{\Main}{\textsc{Main}}

\newcommand{\ModelDescription}{\textsc{ModelDescription}}
\newcommand{\LinkToCode}{\textsc{LinkToCode}}
\newcommand{\Infra}{\textsc{Infra}}
\newcommand{\Runtime}{\textsc{Runtime}}
\newcommand{\Parameters}{\textsc{Parameters}}
\newcommand{\ValidationPerf}{\textsc{ValidationPerf}}
\newcommand{\Metrics}{\textsc{Metrics}}
\newcommand{\NoTrainingEvalRuns}{\textsc{NoTrainingEvalRuns}}
\newcommand{\HyperBound}{\textsc{HyperBound}}
\newcommand{\HyperBestConfig}{\textsc{HyperBestConfig}}
\newcommand{\HyperSearch}{\textsc{HyperSearch}}
\newcommand{\HyperMethod}{\textsc{HyperMethod}}

\newcommand{\ExpectedPerf}{\textsc{ExpectedPerf}}
\newcommand{\Summary}{\textsc{ExpectedPerf}}

\newcommand{\DataStats}{\textsc{DataStats}}
\newcommand{\DataSplit}{\textsc{DataSplit}}
\newcommand{\DataProcessing}{\textsc{DataProcessing}}
\newcommand{\DataDownload}{\textsc{DataDownload}}
\newcommand{\NewData}{\textsc{NewDataDescription}}
\newcommand{\DataLanguages}{\textsc{DataLanguages}}

\newcommand{\Accept}{\textsc{Accept}}
\newcommand{\Track}{\textsc{Track}}
\newcommand{\OverallRec}{\textsc{AvgRec}}
\newcommand{\Reprod}{\textsc{AvgReprod}}
\newcommand{\ChecklistFeedback}{\textsc{ChecklistFeedback}}

\newcommand{\emnlpTwenty}{\textsc{EMNLP 2020}}
\newcommand{\emnlpTwentyOne}{\textsc{EMNLP 2021}}
\newcommand{\naacl}{\textsc{NAACL 2021}}
\newcommand{\acl}{\textsc{ACL 2021}}
\newcommand{\TheConfs}{\textsc{EMNLP 2020} and \textsc{2021}, \naacl{}, and \acl{}}
\newcommand{\TwoConfs}{\textsc{NAACL} and \textsc{ACL 2021}}

\title{Reproducibility in NLP: What Have We Learned from the Checklist? 
}

\author{
  \textbf{Ian Magnusson}$^\spadesuit$ \quad
  \textbf{Noah A. Smith}$^{\spadesuit\diamondsuit}$ \quad
  \textbf{Jesse Dodge}$^\spadesuit$ \\
  $^\spadesuit$Allen Institute for Artificial Intelligence \\
  $^{\diamondsuit}$Paul G. Allen School of Computer Science \& Engineering,
  University of Washington \\
   {\tt \{ianm,noah,jessed\}@allenai.org}
}
\begin{document}
\maketitle

\raggedbottom

\begin{abstract}
Scientific progress in NLP rests on the reproducibility of researchers' claims.
The *CL conferences created the NLP Reproducibility Checklist in 2020 to be completed by authors at submission to remind them of key information to include.
We provide the first analysis of the Checklist by examining 10,405 anonymous responses to it.
First, we find evidence of an increase in reporting of information on efficiency, validation performance, summary statistics, and hyperparameters after the Checklist's introduction.
% submissions reporting checklist items are more often accepted and perceived as reproducible by *CL reviewers.
Further, we show acceptance rate grows for submissions with more \Yes\ responses.
We find that the $44\%$ of submissions that gather new data are $5\%$ \emph{less} likely to be accepted than those that did not; the average reviewer-rated reproducibility of these submissions is also $2\%$ lower relative to the rest.
We find that only $46\%$ of submissions claim to open-source their code, though submissions that do have $8\%$ higher reproducibility score relative to those that do not, the most for any item. We discuss what can be inferred about the state of reproducibility in NLP, and provide a set of recommendations for future conferences, including: a) allowing submitting code and appendices one week after the deadline, and b) measuring dataset reproducibility by a checklist of data collection practices.

\end{abstract}

\section{Introduction}
\begin{figure}[t]
\includegraphics[width=\linewidth]{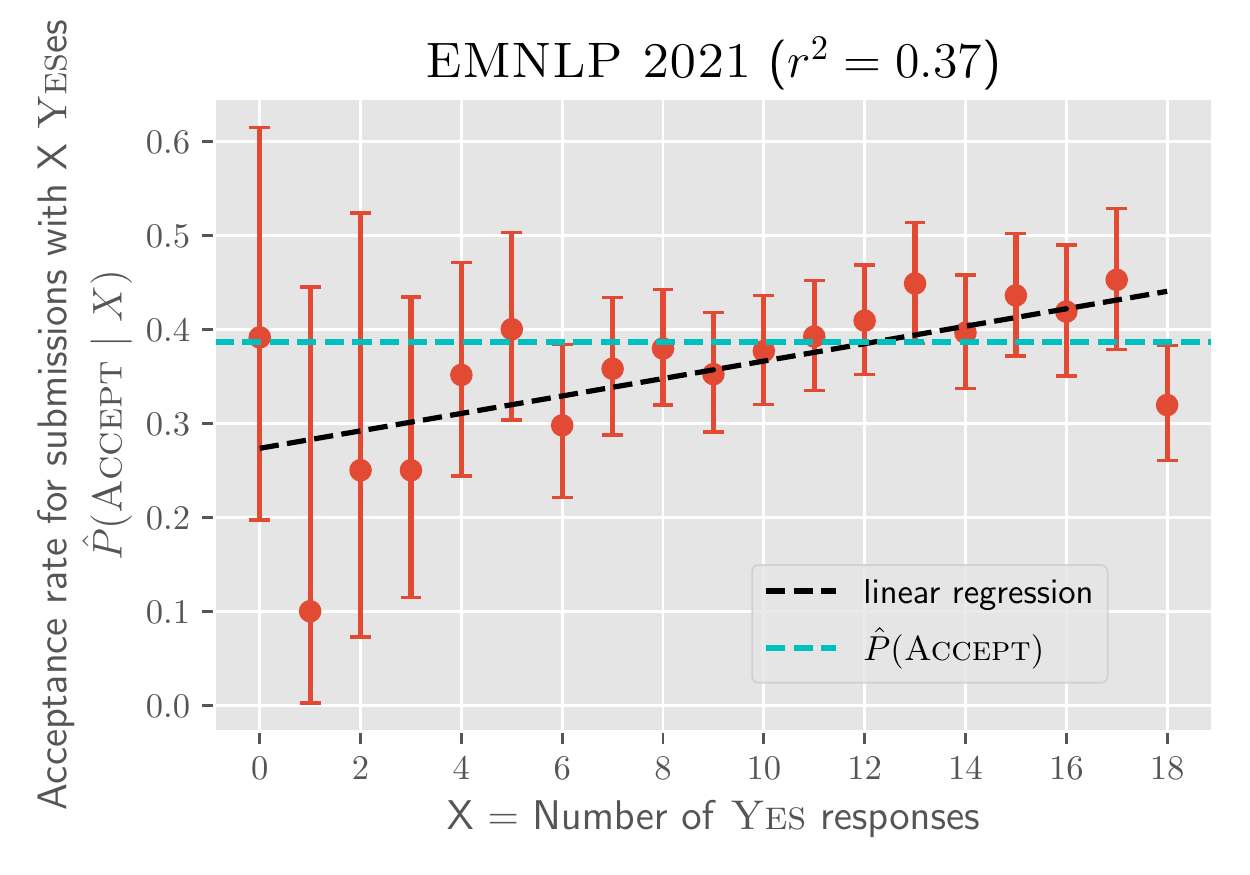}
\centering
\caption{Submissions to \emnlpTwentyOne\ binned by count of \Yes\ responses to the NLP Reproducibility Checklist items. The \Accept\ rate is given for each bin. Papers with more \Yes\ responses are more likely to be accepted, except those that mark \Yes\ to all checklist items, which we hypothesize contain responses which do not accurately represent the associated paper.}
\label{fig:yes_count_to_accept_overall}
\end{figure}

Reproducibility is a foundational component of scientific progress.
NLP systems are complex, and even when their behavior is carefully measured, incentives to publish quickly and limitations in the publishing process can lead to underreporting of information necessary for reproducible science.
%The behavior of NLP systems may be carefully measured by experimenters, but it is also important to track and communicate the experimental decisions that reproduce the behavior.
The ramifications of this extend beyond the research community; the audience of NLP papers published years ago was largely other NLP researchers, but today the world is watching developments in the field, looking for advances that will lead to broadly-adopted applications. As the impact of NLP grows, so too do the consequences of reproducibility in our field.
%No longer is the audience for our research just other NLP researchers, now industry and the public are regularly following cutting-edge research, looking for developments that could be built into the next broadly-adopted NLP system, and thus not only does the reproducibility of our experiments impact the research community, it impacts the world.

%Work on the metascience of reproducibility beyond NLP has explored means of measuring and mitigating what some have even described a ``reproducibility crisis'' in science \cite{doi:10.1126/science.aac4716,Baker20161500SL}.
Of course, NLP is not the first field to evaluate reproducibility; some have even described a ``reproducibility crisis'' in science \cite{doi:10.1126/science.aac4716,Baker20161500SL}.
%Metascience of reproducibility provides a number of interventions for conference organizing.
% Some of the most impactful mechanisms for improving reproducibility lead to increasing reporting and transparency in scientific publications.
% For example, in the computational sciences, conferences can promote publications that also release code, release trained models, or report thorough information about experimental design.
One tool designed to improve reproducibility is a checklist filled out at paper submission time.
% Typically such a checklist includes a list of relevant information that authors could forget to include in their paper;
Such a checklist can descriptively remind authors of relevant information to report, while preserving the freedom for authors to do so however they see fit.
%which may poorly fit some research.
For example, the journal \emph{Nature} requires authors fill out a Reporting Checklist for Life Sciences Articles \cite{ChecklistsNature2018}.
% , which includes subsections for discipline-specific items (like releasing code for computational papers, and reporting population characteristics for studies involving human participants).
In 2019 NeurIPS started to require that submissions fill out the ML Reproducibility Checklist \cite{Pineau2021ImprovingRI}, partly inspired by the \emph{Nature} checklist, and in 2021 AAAI required their own checklist.\footnote{\url{aaai.org/Conferences/AAAI-21/aaai21call/}} CVPR, ICCV, ECCV, and IJCAI provide a checklist but do not collect responses.
%In other fields, checklists have also been required, such as Nature's Reporting Checklist for Life Sciences Articles \cite{ChecklistsNature2018}.

In this work, we provide the first analysis of the NLP Reproducibility Checklist \cite{dodge-etal-2019-show}.
We have gathered 10,405 anonymized responses from \TheConfs. For the latter two we are also able to obtain reviewer scores, reproducibility judgements, and feedback on the Checklist.
% We find that the introduction of the checklist led to an increase in reporting of important information, and ...\jdcomment{other findings here}.
% Our analysis examines:
% \begin{enumerate}
%     \item which items are already commonly reported and selected for by the review process,
%     \item what challenges data collection face in reproducibility,
%     \item how open sourcing code impacts perceived reproducibility and acceptance rates,
%     \item what impact the Checklist had and where can it still improve.
% \end{enumerate}

Our findings include:
\begin{enumerate*}[(1)]
    \item Most checklist items are frequently reported, and submissions reporting them are more often accepted and perceived as reproducible.
    \item Submissions that collect new data are accepted less and viewed as less reproducible, and these gaps are not explained by non-reporting of any current Checklist items. 
    \item Only about half of submissions report open sourcing code and many that do not also lack reporting on efficiency measures and even evaluation metrics.
    \item A majority of reviewers describe the checklist as useful, and by contrasting responses to observed rates prior to the Checklist we evidence a possible increase in reporting.
\end{enumerate*}
We conclude with a discussion of what can be inferred from these findings about the state of reproducibility in NLP and offer recommendations to address the gaps we have measured.

\section{The NLP Reproducibility Checklist}

The NLP Reproducibility Checklist was originally introduced by \citet{dodge-etal-2019-show}. 
Each item on the checklist is phrased as a statement, like ``The number of parameters in each model,'' and authors can mark \Yes\ if they include that information in their paper, \No\ if they do not include it in their paper, or \NA\ if that information does not make sense for their submission (e.g., they do not use any models to report parameter counts for).
The checklist items were a part of the submission form, and it was required that authors fill it out to submit their paper.\footnote{\naacl\ permitted authors to leave items \Blank.} 
Thus, the checklist responses act as a (self-reported) overview of the contents of papers submitted to NLP conferences.
Importantly, authors were not required to include any information in their papers, they were only required to indicate whether or not they did include information.
Answers were made available to reviewers, who were expressly asked to assess the reproducibility of the work. 
% the reviewers seeing the feedback and assessing reproducibility is only mentioned in a FAQ in ACL2021 rather than as part of the checklist instructions.
%The checklist was also presented as a reminder to help authors improve the reproducibility of papers. 
The filled checklists were not released with the published papers.

%With the help of conference organizers, we introduced a reproducibility checklist at premiere NLP conferences. This began with EMNLP 2020 and now includes all *CL conferences. Analysis of the results from these checklists can inform future efforts to encourage reproducibility.

%In this paper we present results and analysis of checklist data from \TheConfs. We obtain data on the checklist results and the publication outcomes of submissions (i.e., acceptance or rejection). This allows us to:
%\begin{itemize}
%    \item Examine the reported frequency of reproducibility practices being used in the NLP community 
%    \item View temporal trends over a one year period
%    \item Identify top patterns of reproducibility practices that co-occur
%    \item Contrast overall acceptance rates with acceptance rates for submissions reporting specific reproducibility practices 
%\end{itemize}
%For some conferences we obtain additional data, including reviewer scores, reviewer reproducibility judgements, and reviewer feedback on the helpfulness of the reproducibility checklist.

%\input{Methodology.tex}
\section{Data and Methodology}

\begin{table}
\footnotesize
\centering 
% \begin{tabular}{lp{1.8cm}p{4.875cm}}
\begin{tabular}{lp{1.8cm}p{4.6cm}}
\toprule
% &{\small Abbreviation} & {\small Full Checklist Item} \\ 
&Abbreviation & Full Checklist Item \\ 
\midrule
\multirow{15}{*}{\rotatebox[]{90}{All Results}}
% \multicolumn{2}{c}{\textit{For all reported experimental results:}}\\
% \midrule
& \tiny \ModelDescription &                                                                                A clear description of the mathematical setting, algorithm, and/or model \\
\cmidrule{2-3}
& \tiny     \LinkToCode &                                              A link to a downloadable source code, with specification of all dependencies, including external libraries \\
\cmidrule{2-3}
& \tiny             \Infra &                                                                                                          A description of computing infrastructure used \\
\cmidrule{2-3}
& \tiny           \Runtime &                                                                                                                       Average runtime for each approach \\
\cmidrule{2-3}
&\tiny         \Parameters &                                                                                                                  The number of parameters in each model \\
\cmidrule{2-3}
& \tiny    \ValidationPerf &                                                                                      Corresponding validation performance for each reported test result \\
\cmidrule{2-3}
& \tiny           \Metrics &                                                                                              Explanation of evaluation metrics used, with links to code \\

\midrule
\multirow{15}{*}{\rotatebox[]{90}{Multiple Experiments}} 
% \multicolumn{2}{c}{\textit{For all results involving multiple experiments, such as hyperparameter search:}} \\
% \midrule

&\tiny    \NoTrainingEvalRuns &                                                                                                        The exact number of training and evaluation runs \\
\cmidrule{2-3}
& \tiny       \HyperBound &                                                                                                                          Bounds for each hyperparameter \\
\cmidrule{2-3}
& \tiny   \HyperBestConfig &                                                                                                Hyperparameter configurations for best-performing models \\
\cmidrule{2-3}
& \tiny       \HyperSearch &                                                                                                                  Number of hyperparameter search trials \\
\cmidrule{2-3}
& \tiny       \HyperMethod & The method of choosing hyperparameter values (e.g., uniform sampling, manual tuning, etc.) and the criterion used to select among them (e.g., accuracy) \\
\cmidrule{2-3}
&  \tiny     \ExpectedPerf &                                                                              Summary statistics of the results (e.g., mean, variance, error bars, etc.) \\

\midrule
\multirow{14}{*}{\rotatebox[]{90}{Datasets}} 
% \multicolumn{2}{c}{\textit{For all datasets used:}} \\
% \midrule

& \tiny        \DataStats &                                                                                                          Relevant statistics such as number of examples \\
\cmidrule{2-3}
&  \tiny        \DataSplit &                                                                                                                 Details of train/validation/test splits \\
\cmidrule{2-3}
 & \tiny   \DataProcessing &                                                                                Explanation of any data that were excluded, and all pre-processing steps \\
 \cmidrule{2-3}
&   \tiny    \DataDownload &                                                                                                            A link to a downloadable version of the data \\
\cmidrule{2-3}
 &  \tiny        \NewData &       For new data collected, a complete description of the data collection process, such as instructions to annotators and methods for quality control \\
 \cmidrule{2-3}
& \tiny     \DataLanguages &                                                                                                  For natural language data, the name of the language(s) \\
\bottomrule
\end{tabular}
\caption{Checklist abbreviations and standardized phrasing. Phrasing per conference in Table~\ref{tab:exact_questions_table} (Appendix).}
\label{tab:questions_summary_table}
\end{table}
\begin{table}[!htbp]
\small
\centering 
\begin{tabular}{lrrrr}
\toprule
Conference & Sub & Wdrn & \multicolumn{2}{c}{\Main/\Findings} \\
\midrule 

%  \emnlpTwenty &
%   3666 &
%   660 \\
%  \emnlpTwentyOne &
%   4815 &
%   1555 \\
%  \naacl &
%   1797 &
%   565 \\
%  \acl &
%   3377 &
%   470 \\
% \midrule
%  Overall &
%   13655 &
%   3250 \\
  \emnlpTwenty &
  3,666 &
  660 &
  \cent{24.92} &
  \cent{14.80}\\
\emnlpTwentyOne &
  4,815 &
  1,555 &
  \cent{25.80} &
  \cent{12.85} \\
\naacl &
  1,797 &
  565 &
  \cent{38.72} &
  N/A\\
\acl &
  3,377 &
  470 &
  \cent{24.42} &
  \cent{15.72}\\
  \midrule
  Overall &
  13,655 &
  3,250 &
  \multicolumn{2}{c}{\cent{39.38}}\\

\bottomrule 
\end{tabular} 
\caption{\textbf{Sub}missions, \textbf{W}ith\textbf{dr}aw\textbf{n}/Desk-Rejects,  and \Main\ conference and \Findings\ acceptance rates in our data.}
\label{tab:data_stats}
\end{table}

In Table~\ref{tab:questions_summary_table} we list the Checklist items and the abbreviations we will use for them throughout this paper. There are three categories of items:
\begin{enumerate*}[(1)]
    \item for all reported experimental results,
    \item for results involving multiple experiments, like hyperparameter search, and
    \item for all datasets used.
\end{enumerate*}
16 of 19 items appeared in all four conferences.
We compare specific checklist items between the ML Reproducibility Checklist and the NLP Reproducibility Checklist conference variations in Appendix~\ref{app:checklist_item_comparisons}.
Full phrasing for each conference is listed in Table~\ref{tab:exact_questions_table} (Appendix).

%\subsection{Form and Distribution}
%The reproducibility checklist was included as part of the submission process for the conferences studied in this analysis. The checklist was implemented as a web form in the respective conference software. Authors were not required to meet all criteria on the checklist, but were asked to fill out the checklist to record which applicable reproducibility practices they performed. In \naacl\ authors did not have to fill out responses in order to move beyond the checklist page, leading to blank responses. In all other conferences participants had to fill in all check list items with a response. Answers were made available to reviewers, who were expressly asked to assess the reproducibility of the work. 
% the reviewers seeing the feedback and assessing reproducibility is only mentioned in a FAQ in ACL2021 rather than as part of the checklist instructions.
%The checklist was also presented as a reminder to help authors improve the reproducibility of papers. 
%The filled checklists were not released with the published paper.

Our data includes, for a given submission, the checklist responses (\Yes, \No, \NA\ for each item), \Main\ + \Findings\ acceptance status (\Accept\ $\in$ \{accepted, rejected\}), and the \Track. No data includes any deanonymizing information, such as authors or paper titles. For \naacl\ and \acl, we have the following metadata for each review: overall recommendation score (``Should this paper be accepted to <conference name>?'') averaged to \OverallRec\ $\in [1,5]$, perceived reproducibility score (``How do you rate the paper’s reproducibility? Will members of the ACL community be able to reproduce or verify the results in this paper?'') averaged to \Reprod\ $\in [1,5]$ or \NA\ if any reviewer responds \NA, and reproducibility checklist feedback (``Are the authors’ answers to the Reproducibility Checklist useful for evaluating the submission?'') aggregated by majority vote to \ChecklistFeedback\ $\in$ \{Not useful, Somewhat useful, Very useful\}.\footnote{Full reviewer instructions available at \url{2021.naacl.org/downloads/NAACL2021-Review-Form.pdf} and \url{2021.aclweb.org/downloads/Review_Form.pdf}}

As shown in Table~\ref{tab:data_stats}, there were a total of 13,655 submissions across the four conferences. 
We remove all withdrawn and desk-rejected submissions from analysis,\footnote{Withdrawn and desk-rejected submissions lack reviews and include blank test submissions and place holders.} comprising $3,250$ submissions ($\cent{23.80}$ of the data), leaving a total of $10,405$ submissions for analysis.

We recognize that the checklist responses are self-reported information, and thus in some cases might not be accurate representations of the associated submission (e.g., authors may mark \Yes\ to an item on the checklist when in fact their paper does not include that information).
We discuss this in Appendix~\ref{app:bad_faith_responses}.

To the best of our knowledge, the creators of the checklist indicated that the data would not be made public; while we currently do not plan to fully open source the data, the data can be made available upon request (and we welcome feedback on this policy).\footnote{``The filled-out checklist will not be released with the published version of an accepted paper, it is meant as a tool for authors and reviewers,'' \cite{dodge_smith_2020}.} 

In all analyses, error bars represent 95\% confidence intervals. These are computed by Clopper–Pearson interval for binary values and bootstrap for continuous values, both using \texttt{scipy} version 1.9.1 \cite{2020SciPy-NMeth}. All comparisons of differences in results are absolute differences unless explicitly stated as relative.

%We recognize that rejected papers have a substantial chance to become part of scientific discourse due to the stochastic nature of reviewing.

%Additionally, two submissions are removed for missing values.

%\input{Results.tex}
% \section{Results}
% In this section we start by exploring correlations between the reproducibility checklist items, the acceptance rate, and the reviewers' estimate of how reproducible a paper is. We then analyze the checklist responses relating to creating and using data, one of the most important topics in NLP. Next, we evaluate checklist items relating to releasing code, which is a major driving factor of reproducibility in any computational science. Finally, we explore how effective the checklist is itself.

%%% figures %%%
% Full figure across conferences
% \begin{figure*}[t]
% \caption{Submissions binned by number of \Yes\ answers. The \Main\ + \Findings\ acceptance rate is given for each bin. \imcomment{We are considering excluding submissions with all identical answers (e.g. furthest right point in these figures) because these likely include bad faith answers (see \ref{tab:all_same_answer_stats}).}
% %Submissions with all answers identical are filtered.
% }
% \label{fig:yes_count_to_accept_ratio}
% \includegraphics[width=\textwidth]{figures/yes_count_to_accept_ratio.png}
% \centering
% \end{figure*}

\begin{figure*}[t]
\includegraphics[width=\textwidth]{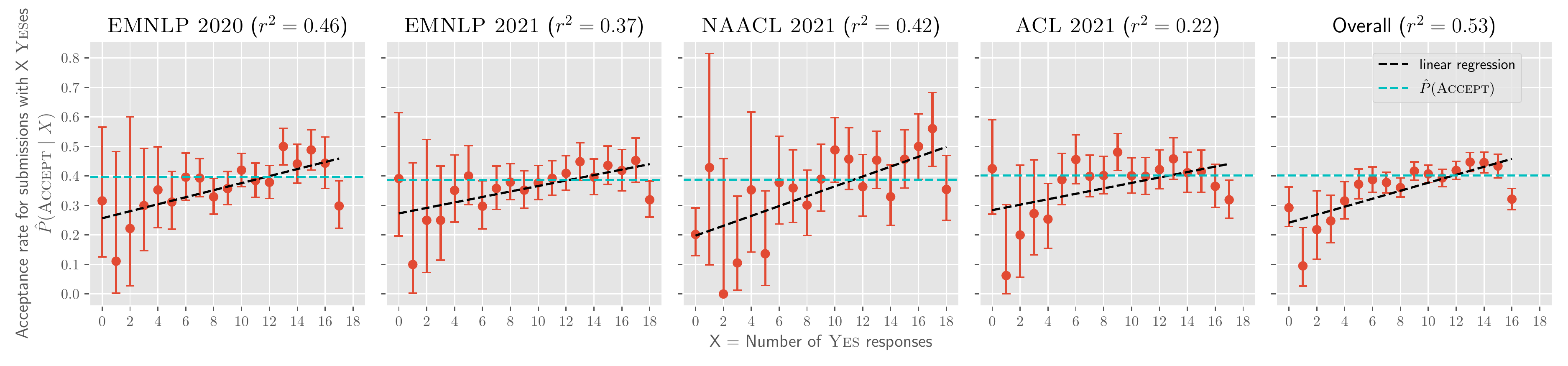}
\caption{\Accept\ rate among submissions binned by count of \Yes\ responses. \Yes\ response count and \Accept\ rate trend consistently positive. All-\Yes\ responses are notably below trend, as discussed in Appendix \ref{app:bad_faith_responses}.
}
\label{fig:yes_count_to_accept_per_conf}
\end{figure*}

\begin{figure}[t]

\includegraphics[width=\linewidth]{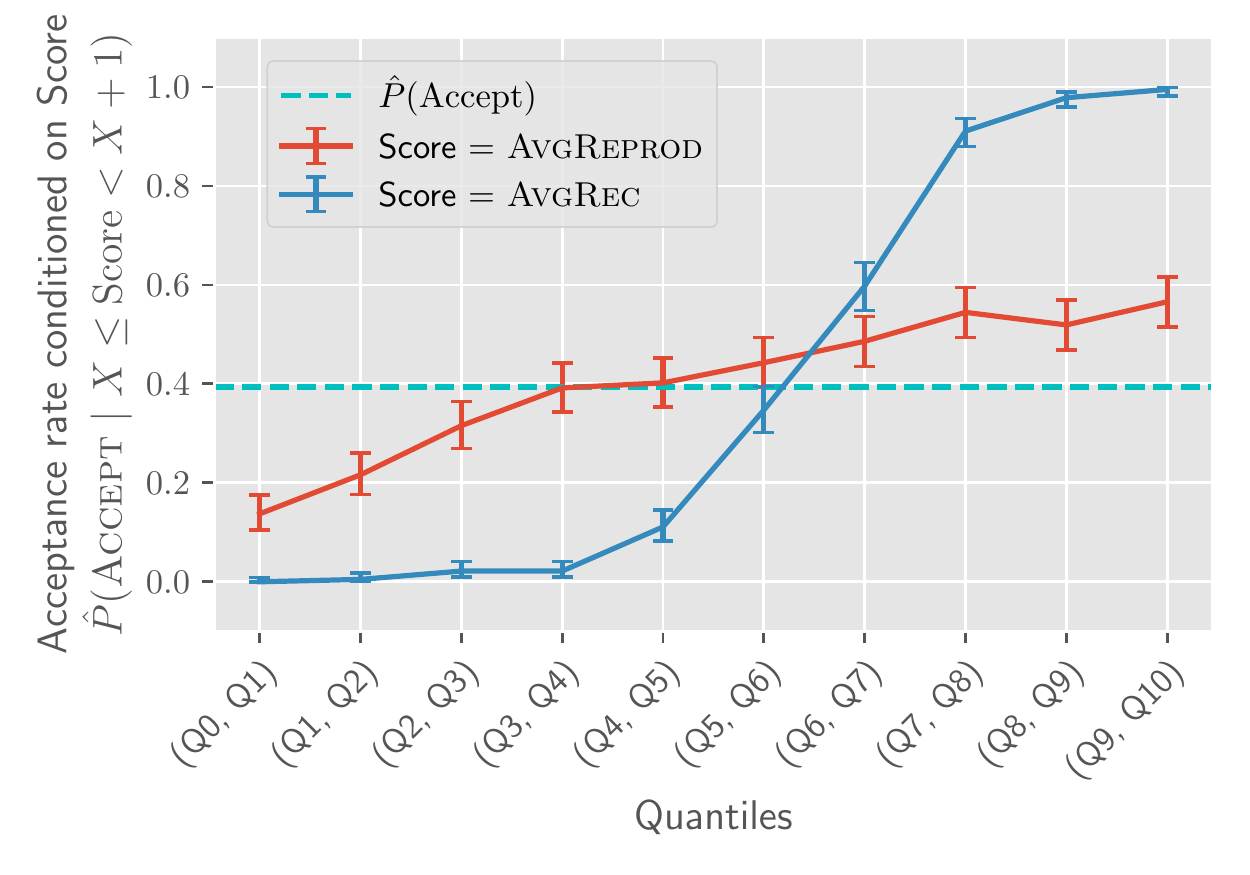}
% \caption{For \naacl\ and \acl\ reviewers rate both Reproducibility and Overall Recommendation. Both associate strongly with acceptance rate ($r^2 = \float{0.76}$ and $r^2 = \float{0.87}$ respectively). Points represent acceptance rate among papers whose average reviewer score rounds to the given $X$ value.}
\caption{\Accept\ rates across quantiles for perceived reproducibility (\Reprod) and overall recommendation (\OverallRec) for \TwoConfs. Perceived reproducibility trends positively with acceptance.}
\label{fig:reviewer_score_to_accept}
\end{figure}

\begin{figure*}[t]
\includegraphics[width=\textwidth]{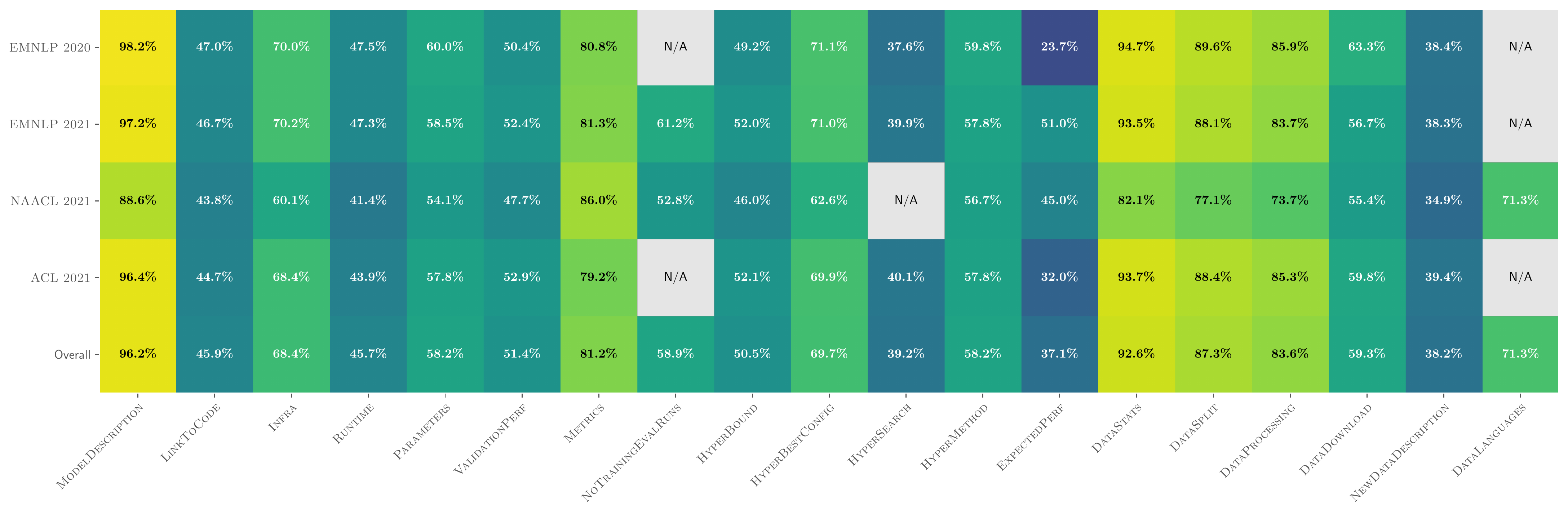}
\caption{\Yes\ response rate per item. Most items are reported for most submissions. Note that \naacl\ respondents were able to leave questions \Blank.
%; These are still counted in the total responses for these ratios. 
Other answers shown in Figure~\ref{fig:portions_all_ans_per_q} (Appendix).}
\label{fig:portions_ans_per_q}
\end{figure*}

\begin{figure}[t]
\includegraphics[width=\linewidth]{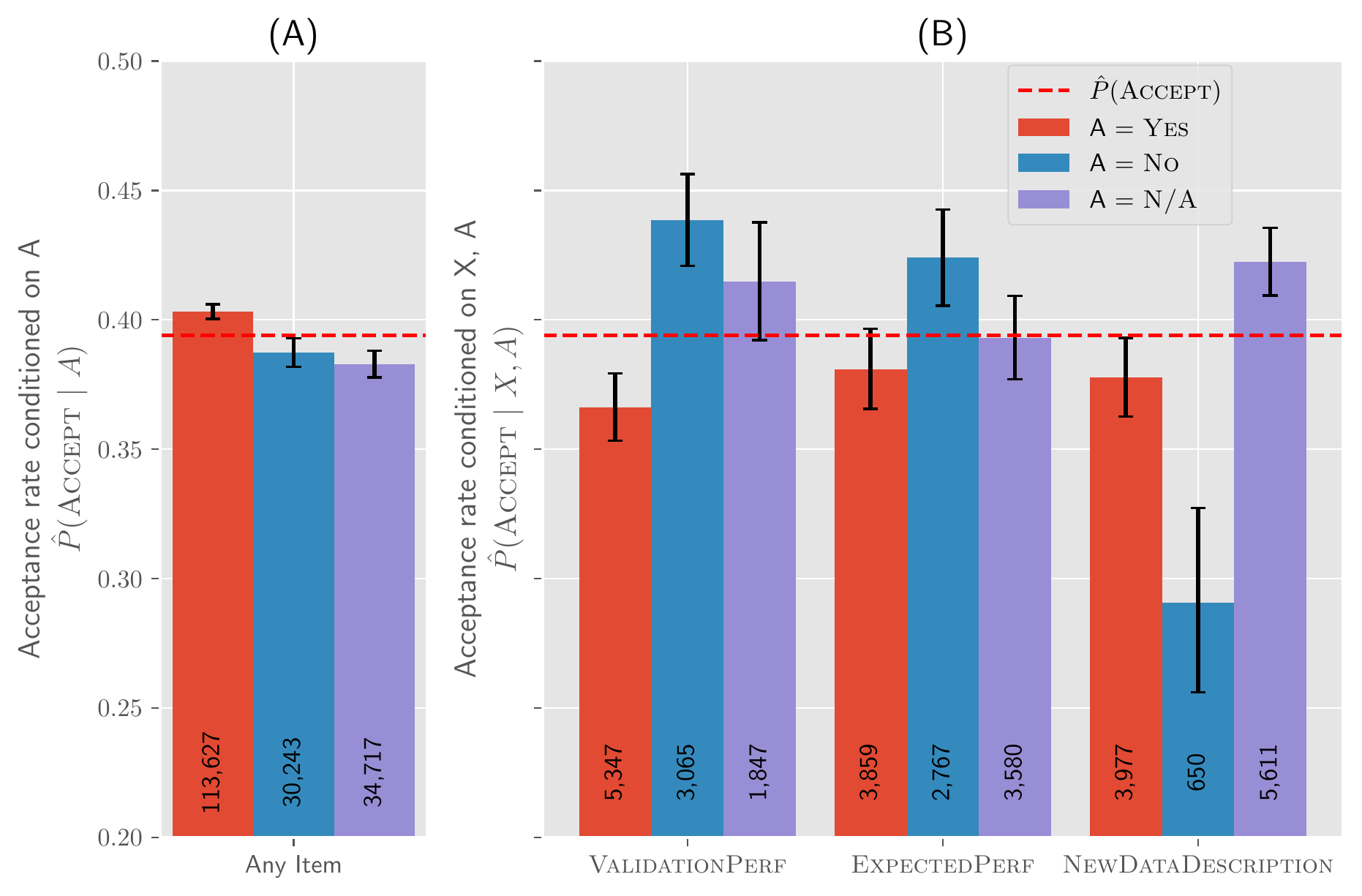}
\caption{\Accept\ rates over all conferences for submissions with a given response. (A) shows rate conditioned on response regardless of item. (B) shows the only items where \Yes\ \Accept\ rates are below average. Total count of each response is shown on the bar. Items with higher \No\ acceptance than \Yes\ could indicate the community has not fully embraced these practices as norms. 
% We would expect more detailed reporting in accepted papers, but items with high \No\ acceptance are actually less common among papers chosen to represent the field through publication. 
High acceptance rate for \NewData\ \NA\  indicates that papers that do not collect data are more likely to be published.}
\label{fig:accept_rates_by_q_overall}
\end{figure}

\begin{figure}[t]
\includegraphics[width=0.95\linewidth]{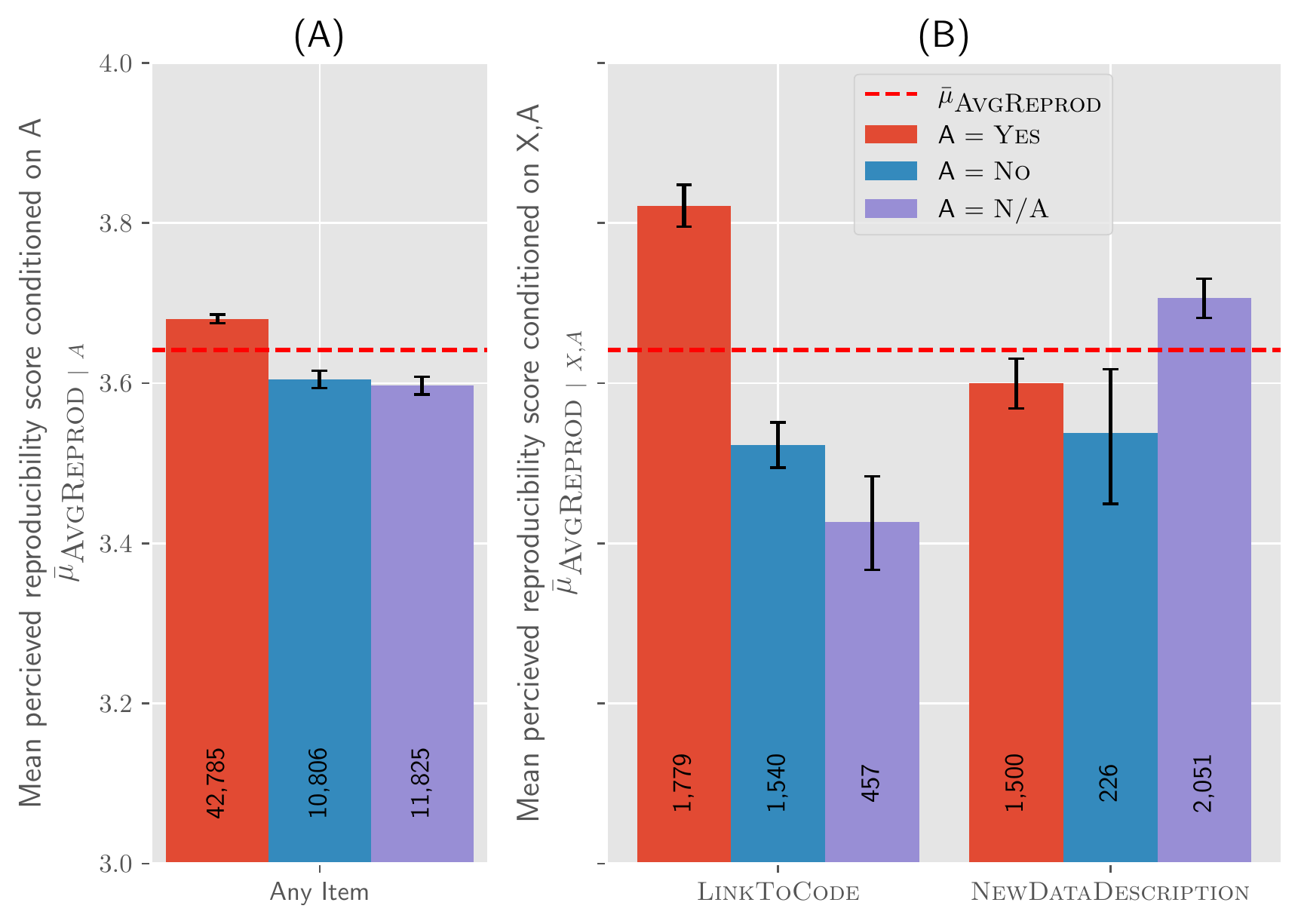}
\caption{Reviewer perceived reproducibility score (\Reprod\ $\in [1,5]$) for submissions with a given response from \TwoConfs, excluding ones with \NA\ \Reprod. (A) shows score conditioned on response regardless of item. (B) shows the items with highest (\LinkToCode) and lowest (\NewData) \Yes\ score. Total count of each response is shown on the bar. \NewData\ is the only item with a below average \Yes\ score.}
\label{fig:reprod_per_answ_overall}
\end{figure}

%%% figures %%%

\section{What Can We Learn About How Reproducibility Already Works?}
We begin by measuring current practice, according to the (self-reported) Checklist data.
%Before presuming anything about reproducibility in NLP, we ought to measure what is already practiced. 
% Although much room for improvement remains, we measure that many checklist items are already regularly reported practices. 
Across all items and conferences, $\cent{62.65}$ of responses were \Yes. Figure~\ref{fig:portions_ans_per_q} shows that most items are reported in most submissions. 
Moreover, we can measure reviewers' perception of reproducibility as well as differences in rates of reporting for items among papers that do and do get accepted by *CL review.
% $\cent{62.65}$

\paragraph{More \Yes\ responses to checklist items associate with higher acceptance.}
In Figures~\ref{fig:yes_count_to_accept_overall} and \ref{fig:yes_count_to_accept_per_conf} we show positive associations between answering more items as \Yes\ and \Accept\ rate. Each point in these figures represents the \Accept\ rate among all the submissions with the same number of \Yes\ responses among items. We regress the \Accept\ rate on a single variable counting checklist items answered \Yes\ for a submission. When pooling responses across all shared questions on all conferences $r^2 = \float{0.53}$.\footnote{The $r^2$ value for these regressions ranges across the conferences from $\float{0.22}$ to $\float{0.46}$, and the trend is consistently positive.} Notably, submissions with \Yes\ responses to all items are consistently below the trend. We hypothesize in 
Appendix~\ref{app:bad_faith_responses} that these submissions include responses which do not accurately represent the associated paper; recall that authors were required to fill out the checklist in order to submit, so marking the same response to all items is, in some sense, as close as they can get to not filling it out. The lower acceptance rate suggests that reviewers are not scoring papers based on the responses to the checklist itself, instead evaluating the contents of the paper, as intended.
%; we hypothesize in Appendix \ref{app:bad_faith_responses} that these submissions include responses which do not accurately represent the associated paper.

\paragraph{Reviewer assessed reproducibility associates with acceptance rate.}
In Figure~\ref{fig:reviewer_score_to_accept} we compare \Accept\ rate across quantiles of \Reprod\ and \OverallRec. Though \Accept\ rate grows much more slowly for \Reprod\ than \OverallRec, reviewers assessment of reproducibility is still evidently associated with acceptance.

\paragraph{In all but three checklist items, \Yes\ responses are associated with higher \Accept\ rates.} Figure~\ref{fig:accept_rates_by_q_overall}.A presents \Accept\ rates conditioned on a given response. \Yes\ responses receive a 
%\nascomment{the percentages in this paragraph have too many sig digits ... just two is fine I think}\imcomment{Jesse and I talked about using .3f for full numbers and \cent{.1} for percents. I'm fine doing it differently but will wait until the final edit to settle this one time.}
$\cent{0.929}$ higher rate than the overall average, while \No\ and \NA\ receive $\cent{0.652}$ and $\cent{1.101}$ lower rates respectively. 
Figure~\ref{fig:accept_rates_by_q_overall}.B presents the three exceptions where answering \Yes\ to that checklist item receives a lower than average rate. These are discussed in detail in Section \ref{sec:dataset_creation} and Appendix \ref{app:additional_results}.

\paragraph{In all but one checklist item, \Yes\ responses are associated with higher \Reprod\ scores.}
Figure~\ref{fig:reprod_per_answ_overall}.A shows the mean \Reprod\ score conditioned on a given response in \naacl\ and \acl. 
Reassuringly, \Yes\ responses receive $\float{0.0388}$ higher scores than average, while \No\ or \NA\ score $\float{0.0369}$ and $\float{0.0445}$ lower than average, respectively. 
Figure~\ref{fig:reprod_per_answ_overall}.B shows that \LinkToCode\ has the highest score, $\float{0.1798}$ above average. 
We also highlight \NewData\ as it is the only item with a lower than average score when answered \Yes. This exception is discussed further in Section \ref{sec:dataset_creation}, where we hypothesize this reflects a lower than average perceived reproducibility of submissions presenting new data.

\section{The Data Collection Gap}
\label{sec:dataset_creation}

\begin{figure}[t]
\centering
\includegraphics[width=0.9\linewidth]{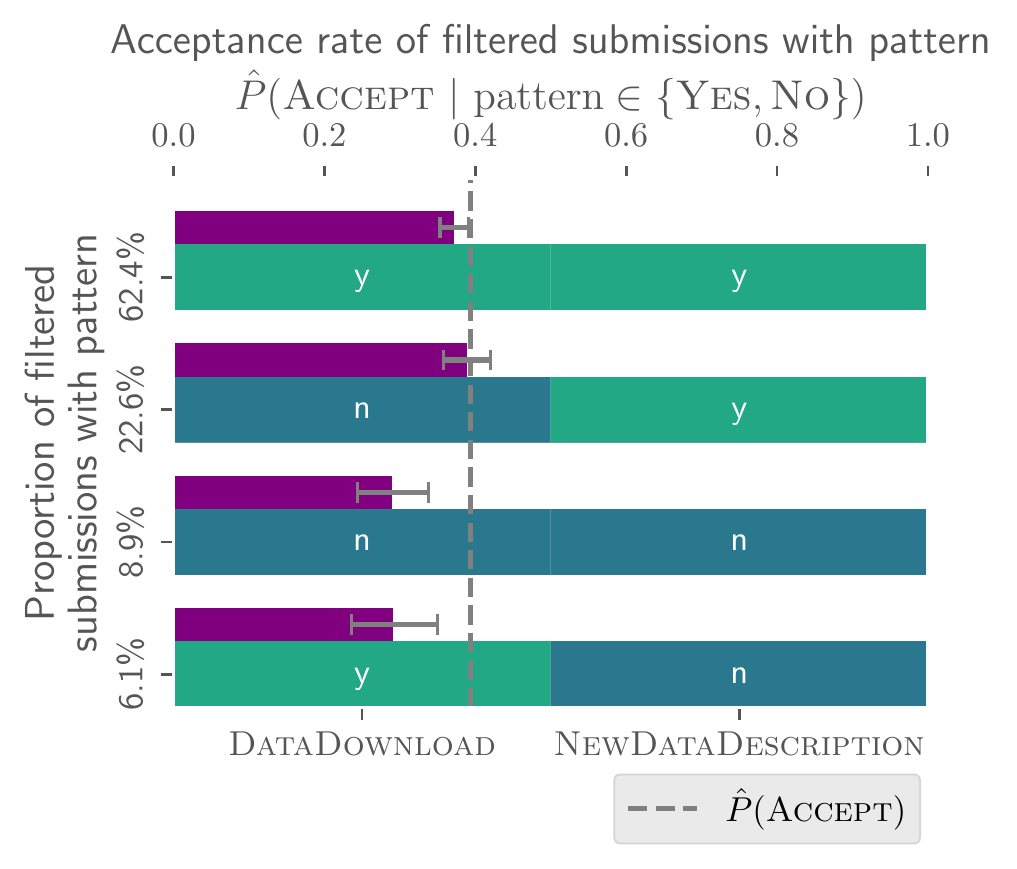}
\caption{Proportion (row labels) and \Accept\ rates (horizontal purple bars) for all response patterns on dataset availability and creation (excluding instances where either item is \NA\ or \Blank). Nearly 1 in 11 of these neither share the data nor describe its collection, yet $\cent{28.91}$ of those are accepted. }
\label{fig:data_creation_and_download}
\end{figure}

Natural language processing has long been a field driven by data. 
% In this section we analyze the checklist items relating to creating or using data.
A body of work has proposed best practices for documenting the characteristics and creation of datasets \cite{Bender2018DataSF, Gebru2018DatasheetsFD, Hutchinson2020TowardsAF, dodge-etal-2021-documenting, rogers-etal-2021-just-think, PushkarnaDatacards2022}.
Among other concerns, such documentation is critical for the difficult task of dataset reproduction \citep[][\emph{inter alia}]{pmlr-v97-recht19a}.
From the checklist item \NewData, which asks that collection is described if new data is presented, we find that $\cent{38.22}$ and $\cent{6.25}$ of submissions mark \Yes\ and \No, respectively.
This implies that $\cent{44.47}$ of submissions to our NLP conferences collect new data; if almost half of submissions collect new data, we argue that data collection and dissemination practices deserve further attention.
This also highlights clear room for improvement in the community: $\cent{14.05}$ of submissions that collect new data do not describe how it was collected, totalling $650$ papers.
%having collected  \jdcomment{X and Y\% of submissions} $\cent{44.47}$ of submissions to the four conferences in our data answer \Yes\ or \No\ to \NewData, which asks for a description of data collection, implying that a submission presents newly collected data. 
%From the checklist responses we see that this accounts for $\cent{44.47}$ of responses across all conferences. 
%While the \No\ responses that do not describe the data collection process are concerning, accounting for $\cent{14.05}$ of submissions presenting new data, even submissions responding \Yes\ to \NewData\ have lower perceived reproducibility and lower acceptance rates.

% this is as basic as \ModelDescription\ but for data, yet many people don't do it.
% \paragraph{More than two in five submissions introduce new data.} \NewData\ was marked as \Yes\ or \No\ in $\cent{44.47}$ of responses across all conferences, indicating that new data was collected. However $\cent{14.05}$ of these answer \No, implying that the submission presents new data but does not provide a complete description of its collection. It is possible that some respondents misunderstood the questions and marked \No\ because they actually did not collect data. Perhaps with greater certainty, we find that at least $\cent{53.93}$ of submissions do \textit{not} present new data, as they select \NA\ for this question.

\paragraph{Submissions with new data have lower than average \Accept\ rate and \Reprod\ scores.}
Alarmingly, submissions that collect new data (i.e., submissions that mark \Yes\ or \No\ to \NewData) have a $\cent{5.11}$ lower acceptance rate than those that do not (i.e., mark \NA\ or \Blank).
A low acceptance rate for answering \No\ to \NewData\ would, by itself, be encouraging, perhaps indicating that reviewers expect data collection to be well documented. However, Figure~\ref{fig:accept_rates_by_q_overall}B shows that, even when answering \Yes\ to describing the data collection process, \Accept\ rate is $\cent{1.62}$ lower than average.
% Specifically, Figure~\ref{fig:accept_rates_by_q_overall}B shows that submissions answering \Yes\ to \NewData\ have a $\cent{1.62}$ lower \Accept\ rate than average, while submissions answering \NA\ are $\cent{2.85}$ higher.
Meanwhile, there is a similar gap in \Reprod. Scores for submissions that collect data are lower by $\cent{2.41}$ relative to those that do not. Again, this is not limited to submissions that fail to describe the data collection process. Figure~\ref{fig:reprod_per_answ_overall}B reveals that the mean score over submissions that do describe their data collection is $\float{0.0412}$ below the mean of all submissions. 
% Figure~\ref{fig:reprod_per_answ_overall}B reveals with mean score over \Yes\ responses $\float{0.0412}$ below the mean of all submissions and \NA\ $\float{0.0645}$ above. 
When considering only accepted papers, however, the gap in \Reprod\ disappears. The review process ends up with accepted dataset papers with similar \Reprod\ to non-dataset papers, but along the way many more dataset than non-dataset submissions are rejected and those have lower \Reprod. We hypothesize that these phenomena arise both because dataset papers may indeed be more challenging to (re)produce and also because of the persistent (and problematic) tendency to value modeling over data collection \citep{rogers-2021-changing}.

\paragraph{High compliance among the dataset checklist items does not reveal the source of the \Accept\ rate and \Reprod\ gap.} \DataStats, \DataSplit, \DataProcessing, and \DataLanguages\ receive the highest rates of reporting other than \ModelDescription\ and \Metrics. This only grows when looking just at submissions presenting new datasets, reaching $\cent{97.34}$, $\cent{91.66}$, $\cent{91.40}$, and $\cent{86.55}$ respectively, and \NewData\ is also reported in $\cent{85.95}$ of these submissions. Unlike the other dataset items, \DataDownload\ is less frequently reported, but its occurrence and associated \Accept\ rates and \Reprod\ scores are similar whether considering submissions presenting new data or not. This suggests that additional checklist items for data collection should be introduced to measure where this gap in perceived reproducibility is coming from.
% -0.19474539236431676 less reprod for datadownload no when new data, but accept rate is similar to yes for datadownload (both are lower but this is b/c dataset papers are just lower)
% -0.09024291716806143 less reprod for datadownload no for dataset or non dataset papers combined

\paragraph{28.9\% of submissions with new data do not provide a downloadable version of the data.}
More generally, a clear area for improvement is that $\cent{25.28}$ of submissions overall answer \No\ to \DataDownload; providing a link to download a dataset is still important for previously released datasets as it might be ambiguous which version of a dataset was used. But for newly collected data, answering \No\ to \DataDownload\ implies the data is not publicly available at all. Moreover, when \DataDownload\ is \No, the rate of submissions reporting the collection process in \NewData\ drops $\cent{14.18}$. Figure~\ref{fig:data_creation_and_download} further reveals the interaction between \DataDownload\ and \NewData. When a new dataset submission provides neither a description of the data collection process nor access to the data itself, this leaves very little for reviewers to assess, at least with regard to the data contributions of the paper. Yet $\cent{28.91}$ of these papers are accepted.

%Presenting new data with all other data reproducibility practices but not linking to it (or listing data link as N/A) occurs in more than a quarter of submissions presenting new data and receives near average acceptance rates.

\section{Code (Un)availability}

\begin{figure}[t]
\includegraphics[width=\linewidth]{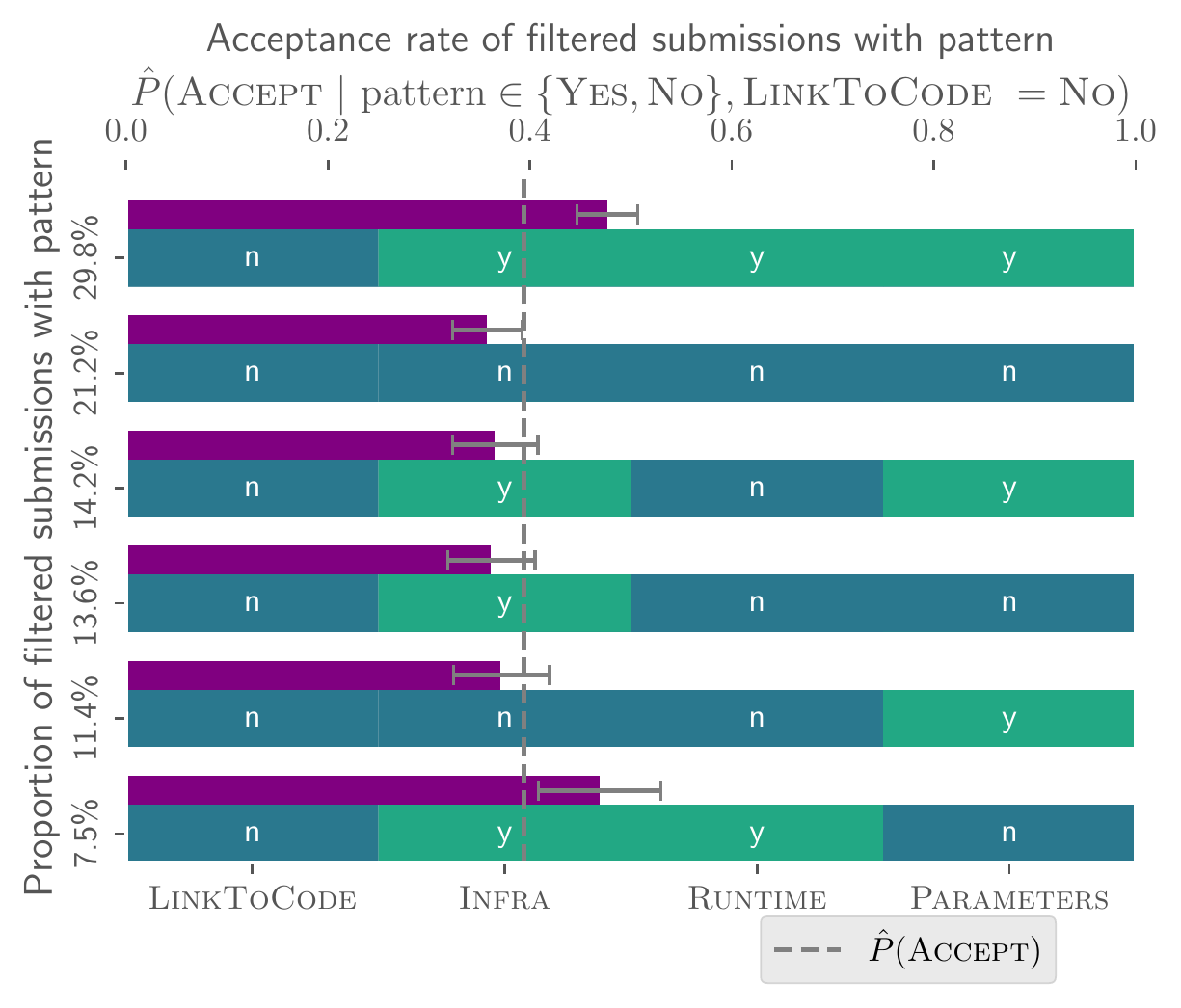}
\caption{Proportion (row labels) and \Accept\ rates  (horizontal purple bars) for efficiency response patterns with $> 100$ submissions when \LinkToCode\ is \No\ and no responses are \NA\ or \Blank. More than 1 in 5 of these do not report any efficiency items, which are difficult to infer without source code.}
\label{fig:efficiency_and_no_link_to_code}
\end{figure}

We see that on average $\cent{45.89}$ of submissions report linking to code ($\cent{47.46}$ for accepted papers).
We see in Figure~\ref{fig:reprod_per_answ_overall} that whether submissions answer \LinkToCode\ as \Yes\ or \No\ has the largest difference in \Reprod\ scores, with a gap of $\float{0.2982}$. 
Yet \Accept\ rates for submissions with or without \LinkToCode\ are nearly the same.

\paragraph{We find similar rates of links to code as at ML conferences.}
 \citet{Pineau2021ImprovingRI} reveal a $\cent{38.76}$ self-reported rate of code availability at submission time for NeurIPS 2019. They find this number drops to $\cent{27.70}$ when checked by at least one reviewer. Extrapolating from this false reporting rate, the true code availability rate among accepted papers in our data might be $\cent{32.80}$. Meanwhile, a study on ICML 2019 by \citet{chaudhuri_salakhutdinov_2019} finds $36\%$ of submitted and $43\%$ of accepted papers have code at submission time, though it unclear if these are self-reported. 

\paragraph{Previous efforts to measure camera-ready code availability have found widely different rates than our reported \LinkToCode\ at submission time.} Unfortunately our data does not cover code availability at camera-ready, except insofar as some authors may interpret this checklist item to permit promises to later release code. $\cent{24.31}$ of papers at NAACL 2022 opted in to submitting a code link to the Reproducibility Track and received an Open Source Code badge.\footnote{\url{naacl2022-reproducibility-track.github.io/}}
We recognize this was optional for authors, and thus it is likely the case that the true number of camera-ready papers that included a link to code was higher.
% but only $\cent{25.62}$ of published papers opt in to this check. 
The studies mentioned before found that $\cent{74.4}$ and $64\%$ of camera-ready papers had links to code at NeurIPS 2019 and ICML 2019. Narrowing the range of these measurements should be a worthwhile effort, as these studies found code being available \textit{during} review was useful in 1,315 reviews in NeurIPS 2019, and $\cent{18.29}$ of ICML 2019 reviewers surveyed were able to look at code and found it useful.

\paragraph{Items on compute efficiency are completely reported in only 29.8\% of submissions without code.} Figure~\ref{fig:efficiency_and_no_link_to_code} shows patterns for these efficiency items that occurred more than 100 times. While \Accept\ rates are somewhat lower when items are not reported, $\cent{21.21}$ of these without-code submissions report none of the efficiency measures. There may be unavoidable impediments to making code available, such as intellectual property. But in this case even greater emphasis should be placed on reporting efficiency measures, as estimating these without code is quite difficult. Similarly $\cent{19.62}$ of submissions with no code report \No\ to explaining \Metrics\, which may render evaluations irrecoverably ambiguous if there are varying implementations of a metric.

\section{How Effective is the Checklist?}

% \nascomment{can we as the question in this section's title without stating what the intent of the checklist was?}\imcomment{I'll add a little intro paragraph stating what the checklist's intent is quoting the blog post.}

\begin{figure}[t]
\includegraphics[width=\linewidth]{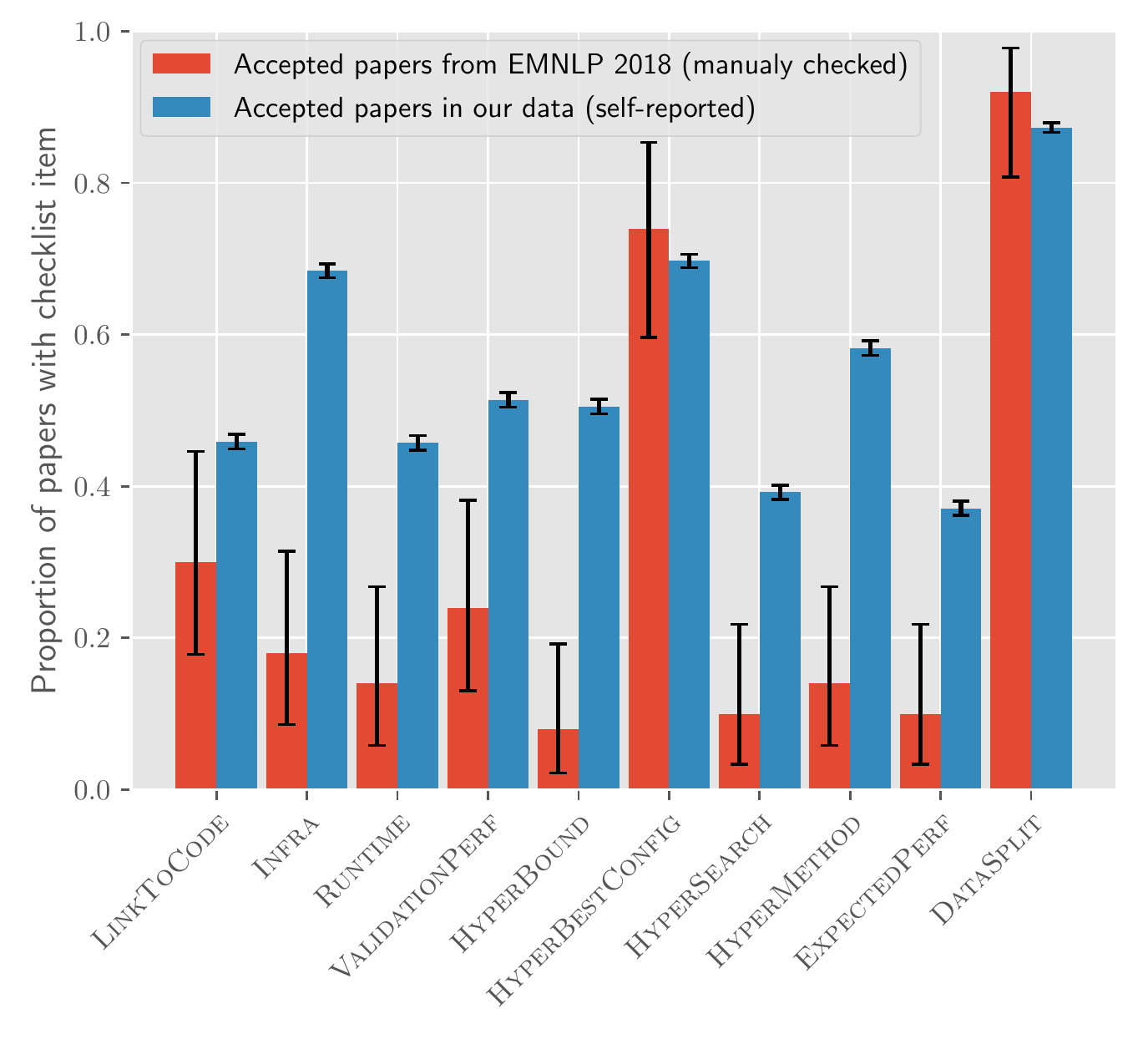}
\caption{Reporting rates before and after the implementation of the Checklist. \citet{dodge-etal-2019-show} manually check a subset of items in 50 randomly sampled EMNLP 2018 papers. We compare to accepted papers from all conferences in our data. While our self-reported data likely overestimate rates, it appears all but 2 items are now reported more often.
}
\label{fig:before_and_after}
\end{figure}

\citet{dodge_smith_2020} describe the Checklist as intended to improve ``reporting of the setup and results of the experiments that authors have conducted.'' Though self-reported data do not directly answer this question, we find potential evidence of such an improvement. Diachronic analysis also shows that reporting rates may have stagnated after initial improvement. We also examine reviewer and author views on the Checklist.

\paragraph{Compared against manually checked data from before the Checklist introduction, our data shows increases in 8 of 10 items.} Figure~\ref{fig:before_and_after} shows rates of a subset of items that were manually checked by \citet{dodge-etal-2019-show} in 50 papers sampled from EMNLP 2018. The self-reported rates available in our data are not ideal comparisons as they likely overestimate. However, the EMNLP 2018 sample may also overestimate, as only ``experimental results'' are included for which we would expect fewer \NA s, given the Checklist's focus on empirical work.

\paragraph{There is little variation in response proportions between conferences.} Excluding two types of outliers likely caused by changes in the Checklist (see Appendix~\ref{app:additional_results}), the maximum difference between conferences for an item is $\cent{6.61}$ and the maximum difference averaged over all items is $\cent{2.21}$. This does demonstrate that measured response patterns are robust across conferences. However it also indicates that reproducibility reporting has stagnated over this one year period.

\paragraph{When asked, a majority of reviewers found the Checklist to be somewhat or very useful.} In \naacl\ and \acl\, reviewers gave feedback on the Checklist. $\cent{59.87}$ found the checklist ``Somewhat Useful,'' $\cent{16.95}$ found it ``Very Useful,'' and $\cent{23.18}$ found it ``Not Useful.'' While this is higher than the $34\%$ of reviewers who answered ``yes'' that the similar NeurIPS 2019 Checklist was ``useful for evaluating the submission'' \cite{Pineau2021ImprovingRI}, it is worth noting that respondents to the question for NeurIPS could answer that they did not read the checklist results.

\paragraph{Author comments from submissions where the majority of reviewers found the Checklist ``Not Useful,'' show possible gaps in checklist coverage.} Some comment on not training models or using hyperparameters from previous work. Many such submissions are represented among the $\cent{22.03}$ that answer \NA\ to all hyperparameter questions.
% This is also visible in the checklist responses, as items involving multiple experiments receive the most \NA\ responses other than \NewData. In particular, $\cent{22.03}$ of submissions answer \NA\ to all shared hyperparameter questions. The overall \Accept\ rate is within the 95\% confidence interval of the \Accept\ rate of these submissions, so adverse effect from the checklist is unlikely. 
Others comment on referring readers to citations for details of standard models, data, or metrics. Re-elaboration is pedagogically important but, comments argue, especially onerous for survey papers. Finally a comment notes that the Checklist is less relevant to psycholinguistics and cognitive modeling, and indeed the \NA\ rate of ``Linguistic Theories, Cognitive Modeling and Psycholinguistics'' \Track\ submissions is $\cent{33.66}$, an increase of $\cent{14.59}$ above the \NA\ rate over all \Track{}s.

\section{Discussion}
% explictly say recommendations both for the checklist and beyond
% something about covering gaps seen in last section
% something about stagnation

Our findings from the NLP Reproducibility Checklist can both help inform new interventions and guide improvements to future checklists that will measure the outcomes of those interventions. These findings suggest that, after an initial increase, rates of reporting have stagnated in the period examined and will need new approaches to improve further.

% We provide recommendations that aim to improve measurement, reporting, and coverage of reproducibility information. \nascomment{can cut this sentence}\imcomment{agreed!}

\paragraph{The conference system should better support papers that collect new data.}
As discussed in Section~\ref{sec:dataset_creation}, papers that collect new data have \cent{5.11} lower acceptance rate than those that do not. Whether or not this gap is a cause or effect of the lack of prestige given to data work that \citet{rogers-2021-changing} describes, increasing awareness and resources for this work can help more high quality data reach publication.
Checklists should also increase coverage of this topic.
In our data a single item, \NewData, covers all reporting regarding data collection. We find that papers with new data are perceived as less reproducible both when answering \No\ or \Yes\ to describing how they collected data. Likely a combination of several factors lead reviewers to score the reproducibility of papers with new data lower by \cent{2.41} relative to papers without. To discover which are lacking, best practices in data reproducibility documentation \cite{Gebru2018DatasheetsFD, dodge-etal-2021-documenting} should be tracked individually with checklists.

\paragraph{Incentivize authors releasing code.}
We find that releasing code is the single most influential checklist item on perceived reproducibility. This aligns with work across diverse fields that argues open source code is key for transparent and reproducible science \cite{eglen2017toward, celi2019plos, shamir2013practices}. These works also suggest that beyond reproducibility, open source code enables more impactful research by allowing other researchers to build on introduced methods and better understand findings through reading code. However, we find that less than half of papers in our study report releasing code  at submission.
We encourage conferences to incentivize code release at submission and especially camera-ready, and authors should be made aware of the significant benefit that code submission can have for the review process.\footnote{Even when code cannot be made publicly available due to intellectual property concerns, private code submission should be facilitated and extra emphasis should be placed on reporting items such as efficiency measures that are hard to reconstruct without public code.}
Initiatives like the NAACL 2022 Reproducibility Track are a step in the right direction, as they publicly recognize open source code and verify code availability rather than only relying on self-reporting.
However, in our data we see no evidence that code availability is increasing over time, so more direct incentives from publication venues are needed.

\paragraph{Make checklist responses public.}
Self-reported data is notoriously unreliable, but making the checklist responses public will add accountability.\footnote{This should still be combined with studies that manually audit reporting in papers.} 
In addition, the checklist responses can reference specific sections and act as an index of the paper, so a reader knows where to look for what information.
This will be implemented at ACL 2023, and we recommend other conferences follow.

\paragraph{Conferences should allow submission of checklists, unlimited appendices, and code a week after the main deadline.}
% Following ICLR \jdcomment{some year}\ianm{I looked an ICLR doesn't seem to have had a later deadline for supplementary materials in the last several years. ICML 2020 seems to have had an extra week for code but it's not clearly documented and is explicitly absent after 2020. So we might not want to emphasize that if they actually stopped doing it.}, We recommend allowing authors to submit code and appendices one week after the main deadline.
Doing so can help establish a norm of code submission as \emph{part} of the review process. Likewise, additional time could improve completeness and accuracy of the checklist.
Many pieces of information important for reproducibility are appropriate to include in the appendix of a paper without counting towards the page limit (e.g., a full list of hyperparameter values). This need not increase the burden on reviewers, as they can consult checklists rather than the appendix to assess reporting. 
%Such information likely would be referenced by someone wanting to build on the work, rather than reviewers initially evaluating it.

\subsection*{Looking Forward}
%Checklists collected during submission can measure practices in NLP at a comprehensive scale, but care must be taken to improve the quality of this self-reported data. Treated with the same methodical approach as the rest of our science...
%Further questions should be developed as standard practices formalize and evolve.

Checklists collected during submission can measure practices in NLP at a comprehensive scale. 
To our knowledge, our work and \citeposs{Pineau2021ImprovingRI} are the only analyses of submitted reproducibility checklists at AI conferences. These are examples of metascience in AI, or applying scientific rigor to the process of AI research; we expect that as NLP matures, we will see more examples of work analyzing and improving the scientific process.
There are also examples of other work which manually audits papers \cite{fokkens-etal-2013-offspring, Gundersen_Kjensmo_2018, McDermottML4Health2019, HaibeKains2020TransparencyAR, marie-etal-2021-scientific}, which can compliment self-reported checklists, and other conference submission metadata \cite{Chen2022ICLR}, with validated samples. 

As standard practices in our field evolve, we will have to update all parts of the conference process, from checklists to reviews to paper presentations.
As a positive example, ACL Rolling Review implemented the Responsible NLP Checklist,\footnote{\url{aclrollingreview.org/responsibleNLPresearch/}} which includes ethics as well as reproducibility items.
While we do not have data with which to evaluate the Responsible NLP checklist, our findings show the need for just such efforts to expand the coverage of checklists to better serve the community.

\section*{Limitations}
Our analyses rely on data from checklists filled in by authors and ratings provided by reviewers. Checklists are self-reported and thus not necessarily accurate. We discuss where these bad faith resposes might influence our results in Appendix~\ref{app:bad_faith_responses}.
% In particular we note that responses answering all items with the same answer are less likely to be accepted. These possibly include responses quickly filling out all the answers as \Yes\ to bypass the checklist. However even if some \Yes\ answers are given in bad faith other responses in the same submission may still be accurate, and it is unknown how many of these all-same submissions have this issue so we retain these for analysis.
Another data limitation is that phrasing changes between conferences for some items, and 3 items do not appear in all conferences (see Appendix~\ref{app:checklist_item_comparisons}). \naacl\ also introduces \Blank\ as a possible answer when respondents do not choose any answer. There is also possible ambiguity between the \No\ and \NA\ answers as it is apparent from the checklist open text comments that some authors used \No\ when the item was not applicable to their work. Our data also only covers four conferences across 2020 and 2021, and as such it is difficult to assess any temporal trends. Reviewer data is also subject to inaccuracy; for instance reviewer perceived reproducibility scores are only subjective estimations of the likelihood of actual reproducibility. Rushed reviewers could easily miss where some important information is reported in a paper. Moreover, we only have reviewer data for 2 of 4 conferences.

Our finding that papers that collect data have a gap in acceptance and perceive reproducibility relies on an indirect inference about which papers collect data. Checklists did not ask this explicitly but rather \NewData\ should be answered \NA\ for all papers that do not collect data. 
% As previously mentioned some authors may use \No\ to indicate non-applicability, and others may be filling in all \Yes\ or \NA\ for expediency regardless of whether or not they collect data. Moreover even assuming accurate data, our analyses do not assert what causes the lower acceptance rate. Even though we find that reviewer perceived reproducibility scores are lower for data collection submissions, we do not know that this contributes to the lower acceptance rate.

Our findings about code and data availability are limited by the ambiguity of when they must be made available to qualify for answering \Yes. It is evident from the open text checklist comments that some authors answer \Yes, \No, or even \NA\ when they have not yet made code or data available but plan to do so on acceptance.

Any self-reported inaccuracies in our data would particularly affect our findings about the impact of the Checklist introduction on reporting rates. By definition, we are not able to compare to self-reported rates from before Checklist introduction, so we instead rely on \citeposs{dodge-etal-2019-show} manually checked rates. 9 of the items in our data are not covered in the previous work, but the items that are share have similar phrasing.

Finally, pooling results over conferences can obscure conference-specific dynamics, such as differences in which items have lower than average \Yes\ \Accept\ rates discussed in Appendix~\ref{app:additional_results}. We check that trends that we highlight in our analyses are consistent across conferences.  And we also present unaggregated figures in the appendix. Likewise, we find that \Accept\ rates are nearly identical across conferences (see Appendix~\ref{app:baseline_acceptance_rates}), enabling us to contrast against an overall acceptance rate.

% misc limitations
% cant share data

% ACL 2023 requires all submissions to have a section titled ``Limitations'', for discussing the limitations of the paper as a complement to the discussion of strengths in the main text. This section should occur after the conclusion, but before the references. It will not count towards the page limit.
% The discussion of limitations is mandatory. Papers without a limitation section will be desk-rejected without review.

% While we are open to different types of limitations, just mentioning that a set of results have been shown for English only probably does not reflect what we expect. 
% Mentioning that the method works mostly for languages with limited morphology, like English, is a much better alternative.
% In addition, limitations such as low scalability to long text, the requirement of large GPU resources, or other things that inspire crucial further investigation are welcome.

\section*{Ethics Statement}
Scientific reproducibility is key to the benefits science can bring to society. Simply put, findings that cannot be reproduced cannot be relied upon, which can lead to wasted societal resources or even to harmfully incorrect understandings that misguide interventions. Our work focuses on the use of checklists to improve reporting of reproducibility information in scientific publications. While overly prescriptive and general rules about reproducibility could stifle less represented research communities whose practices may be less well understood by conference organizers, checklists attempt to mitigate this risk by only reminding authors of possibly salient information while still permitting authors to determine which items are or are not applicable.

At the same time, checklists which are filled out and collected for data analysis have the additional ethical risks associated with work that attempts to make social practices legible. That is, a checklist may neglect to cover practices used in a research community and thereby efface their role in the overall scientific endeavor, or conversely some practice may receive unfair scrutiny in excess of that given to other more prestiged practices. In the long term, checklists are perhaps most important as documents for guiding new generations of researchers writing their first papers, and thus even without being enforced they may still be taken as normative statements about best practices in the field.

To guide efforts to improve reproducibility in the field of NLP, we have analyzed responses to the NLP Reproducibility Checklist collected by four conferences. The Checklist data is covered by the default terms as it has no stated license, and we use it with direct permission from the conference organizers who collected it. The authors of the first version of the checklist state that it is intended for ``improved reporting of the setup and results of the experiments that authors have conducted'' and that it will be used to ``quantitatively analyze our checklist responses'' \cite{dodge_smith_2020}.

We have endeavored to maintain the privacy of respondents by keeping the data anonymized and presenting results at a sufficient level of aggregation to prevent deanonymization. Nevertheless all work that seeks to describe the opinions of groups of humans caries an ethical burden to do so accurately and consistently with the wishes of those represented. To that end, we take care to point out limitations in what can be inferred from the data, and as originally intended by the data creators we do not make the data publicly available.
% we also include full analyeses in the appendix to allow future work to build on these findings.

%Scientific work published at ACL 2023 must comply with the ACL Ethics Policy.\footnote{\url{https://www.aclweb.org/portal/content/acl-code-ethics}} We encourage all authors to include an explicit ethics statement on the broader impact of the work, or other ethical considerations after the conclusion but before the references. The ethics statement will not count toward the page limit (8 pages for long, 4 pages for short papers).

\section*{Acknowledgements}
We thank the organizers and Program Chairs that provided the data for our analysis: Trevor Cohn, Yulan He, Yang Liu (EMNLP 2020), Fei Xia, Wenjie Li, Roberto Navigli (ACL 2021), Anna Rumshisky, Luke Zettlemoyer, Dilek Hakkani-Tur (NAACL 2021), Xuanjing Huang, Lucia Specia, Scott Wen-tau Yih (EMNLP 2021).
We thank Julian Michael, Kyle Lo, Lucy Lu Wang, and Ari Holtzman for fruitful conversations about metascience and feedback on paper drafts. 

% Ari Holtzman - for conversations about survey papers and thoughts on reproducibility of datasets
% Kyle Lo
% Lucy Lu Wang
% Roy Schwartz
% Julian Michael

\bibliography{anthology,custom}
\bibliographystyle{acl_natbib}

\appendix
\section{Appendix}
\label{sec:appendix}

\subsection{Item Comparisons Across Checklists}
\label{app:checklist_item_comparisons}
In order to support comparison, 11 of the 19 checklist items correspond to specific items in the Machine Learning Paper Reproducibility Checklist,\footnote{\url{cs.mcgill.ca/~jpineau/ReproducibilityChecklist.pdf}} a version of which was used at NeurIPS 2019 \cite{Pineau2021ImprovingRI}. A further 4 items are either combinations or decomposition of items from this checklist. 2 more, \ValidationPerf\ and \HyperSearch, are incorporated that are manually evaluated along with 8 items from \citet{Pineau2021ImprovingRI} on a random sample of 50 papers from EMNLP 2018 in \citet{dodge-etal-2019-show}. In that analysis at least one checklist item was found per paper and each checklist item occurred in at least one paper. Finally, \Parameters\ is included for its important role in measuring the complexity of models, and \DataLanguages\ is included because of the importance of acknowledging which communities of speakers are being served by a language technology as noted by \citet{bender_2019}.

The phrasing overlap in NLP and ML Checklists permits comparison of our data to responses from NeurIPS 2019. \citet{Pineau2021ImprovingRI} find similar rates of reporting for dataset and efficiency items, though fewer submissions respond \NA\ to describing data collection. They find higher rates for items concerning hyperparameters and multiple experiments. Most of all their acceptance rates conditioned on items differ dramatically from ours. All but one item for ``empirical results'' get lower than average acceptance for \Yes\ and higher for \NA, while our data shows lower \Yes\ \Accept\ rates for only 3 empirical items. This suggests the applicability of the NLP Checklist is more aligned with reviewing at the studied conferences.

\begin{table}
\caption{Checklist item phrasing differences across conferences. $\Delta$ marks differing item phrasing. \NA\ marks conferences with no equivalent item.}
\label{tab:phrasing_changes}
\tiny
\centering 
\begin{tabular}{lllll}
\toprule
        Abbreviation & \rotatebox[origin=lB]{60}{EMNLP 2020} & \rotatebox[origin=lB]{60}{EMNLP 2021} & \rotatebox[origin=lB]{60}{NAACL 2021} &   \rotatebox[origin=lB]{60}{ACL 2021} \\
\midrule
  \ModelDescription & \checkmark & \checkmark & \checkmark & \checkmark \\
        \LinkToCode & \checkmark &   $\Delta$ &   $\Delta$ &   $\Delta$ \\
             \Infra & \checkmark & \checkmark & \checkmark & \checkmark \\
           \Runtime & \checkmark &   $\Delta$ &   $\Delta$ & \checkmark \\
        \Parameters & \checkmark & \checkmark & \checkmark & \checkmark \\
    \ValidationPerf & \checkmark & \checkmark & \checkmark & \checkmark \\
           \Metrics & \checkmark & \checkmark &   $\Delta$ & \checkmark \\
\NoTrainingEvalRuns &        N/A & \checkmark & \checkmark &        N/A \\
        \HyperBound & \checkmark & \checkmark & \checkmark & \checkmark \\
   \HyperBestConfig & \checkmark & \checkmark & \checkmark & \checkmark \\
       \HyperSearch & \checkmark & \checkmark &        N/A & \checkmark \\
       \HyperMethod & \checkmark & \checkmark &   $\Delta$ & \checkmark \\
      \ExpectedPerf &   $\Delta$ & \checkmark & \checkmark &   $\Delta$ \\
         \DataStats & \checkmark &   $\Delta$ &   $\Delta$ & \checkmark \\
         \DataSplit & \checkmark & \checkmark & \checkmark & \checkmark \\
    \DataProcessing & \checkmark & \checkmark & \checkmark & \checkmark \\
      \DataDownload & \checkmark &   $\Delta$ &   $\Delta$ & \checkmark \\
           \NewData & \checkmark & \checkmark & \checkmark & \checkmark \\
     \DataLanguages &        N/A &        N/A & \checkmark &        N/A \\
\bottomrule
\end{tabular}
\end{table}

The phrasing of the Checklist items was determined by distinct groups of organizers for each conference. While 9 items maintain the same phrasing, 6 see phrasing changes, and 3 are only asked at some conferences (see Tables~\ref{tab:phrasing_changes} and \ref{tab:exact_questions_table}). \LinkToCode\ remains the same in substance while phrasing variations address logistics such as file formats and anonymization. \Runtime\ varies in \emnlpTwentyOne\ and \naacl\ by asking for runtime or energy cost. \Metrics\ varies in \naacl\ by not specifying links to metric code. \ExpectedPerf\ varies in \emnlpTwenty\ and \acl\ by asking for mean \textit{and} variance of hyperparameters, where in the other phrasing any summary statistic of results is sufficient. \DataStats\ includes languages and label distributions in its variations. \DataDownload\ varies only in file formats, except for \naacl\ which also allows for a simulation environment.

\subsection{Bad Faith Responses}
\label{app:bad_faith_responses}
As expected the \ModelDescription\ question was answered \Yes\ by almost all submissions ($\cent{96.26}$ of responses over all three conferences). This question was intended as an attention check and was designed such that almost all submissions should answer \Yes. This helps assure that respondents are not using the \NA\ (or \No) response in protest or bad faith to quickly fill in meaningless answers, as only $\cent{2.60}$ (or $\cent{0.21}$) of submissions answer \ModelDescription\ this way. However this does not preclude the use of answering questions \Yes\ in bad faith to bypass the checklist. Likewise we see $\cent{7.95}$ of \naacl\ respondents leave this field \Blank.

\begin{table*}[htbp]
\caption{Submissions with all Checklist responses given the same answer (e.g., responding \NA\ to all items) and their change in \Main\ and \Findings\ acceptance rate from overall rate.
% All \Yes\ and \Blank\ have notably lower \Main\ conference rates and account for several percent of all submissions.
}
\label{tab:all_same_answer_stats}
\centering 
\begin{tabular}{llrr}
\toprule 
 % &  &  & \multicolumn{2}{c}{$\Delta$ Accept Rate} \\
 % \cmidrule{4-5}
Response & Conference & Submissions & \Accept \\
\midrule 
\parbox[t]{2mm}{\multirow{3}{*}{\Yes}}
& EMNLP 2020 & 134 (\cent{4.46}) & \cent{-9.87}\\
& EMNLP 2021 & 238 (\cent{7.30}) & \cent{-6.72}\\
& NAACL 2021 & 79 (\cent{6.41}) & \cent{-3.27}\\
& ACL 2021 & 213 (\cent{7.33}) & \cent{-8.22}\\
\midrule 
\parbox[t]{2mm}{\multirow{3}{*}{\No}}
& EMNLP 2020 & 1 (\cent{0.03}) & \cent{-39.72}\\
& EMNLP 2021 & 0 (\cent{0.00}) & -\\
& NAACL 2021 & 0 (\cent{0.00}) & -\\
& ACL 2021 & 0 (\cent{0.00}) & -\\
\midrule 
\parbox[t]{2mm}{\multirow{3}{*}{\NA}}
& EMNLP 2020 & 17 (\cent{0.57}) & \cent{-4.43}\\
& EMNLP 2021 & 22 (\cent{0.67}) & \cent{2.26}\\
& NAACL 2021 & 15 (\cent{1.22}) & \cent{14.62}\\
& ACL 2021 & 40 (\cent{1.38}) & \cent{2.36}\\
\midrule 
\Blank
& NAACL 2021 & 89 (\cent{7.22}) & \cent{-24.11}\\
\midrule 
\textbf{All Same}
& \textbf{Overall} & \textbf{848 (\cent{8.15})} & \textbf{-\cent{8.13}}\\

% absolute rates
% \midrule 
% \parbox[t]{2mm}{\multirow{3}{*}{\rotatebox[origin=c]{90}{\Yes}}}
% & EMNLP 2020 & 134 & 17.91\% & 11.94\%\\
% & EMNLP 2021 & 238 & 18.49\% & 13.45\%\\
% & NAACL 2021 & 79 & 35.44\% & N/A\\
% \midrule 
% \parbox[t]{2mm}{\multirow{3}{*}{\rotatebox[origin=c]{90}{\No}}}
% & EMNLP 2020 & 1 & 0.00\% & 0.00\%\\
% & EMNLP 2021 & 0 & 0.00\% & 0.00\%\\
% & NAACL 2021 & 0 & 0.00\% & N/A\\
% \midrule 
% \parbox[t]{2mm}{\multirow{3}{*}{\rotatebox[origin=c]{90}{\NA}}}
% & EMNLP 2020 & 17 & 29.41\% & 5.88\%\\
% & EMNLP 2021 & 22 & 27.27\% & 13.64\%\\
% & NAACL 2021 & 15 & 53.33\% & N/A\\
% \midrule 
% \parbox[t]{2mm}{\multirow{3}{*}{\rotatebox[origin=c]{90}{\Blank}}}
% & EMNLP 2020 & 0 & 0.00\% & 0.00\%\\
% & EMNLP 2021 & 0 & 0.00\% & 0.00\%\\
% & NAACL 2021 & 89 & 14.61\% & N/A\\

\bottomrule 
\end{tabular} 
\end{table*}
Submissions with all identical answers have lower than average acceptance. In Table~\ref{tab:all_same_answer_stats} we show counts and change from average acceptance rates for submissions whose answers are all identical. This pattern is most prevalent for \Yes\ and \Blank, accounting for several percent of all submissions. All \No\ and \NA\ submissions, however, are quite infrequent. One possible explanation is that selecting all \Yes\ or \Blank\ is an expedient way to bypass the checklist during the submission process. Though we cannot know what portion of submissions with this pattern may exhibit this issue, it is important to be aware of this limitation.

\subsection{Additional results}
\label{app:additional_results}

\begin{figure*}[t]
\caption{The portion of submissions giving a particular response per question. Note that \naacl\ respondents were able to leave questions \Blank; These are still counted in the total responses for these ratios.}
\label{fig:portions_all_ans_per_q}
\includegraphics[width=\textwidth]{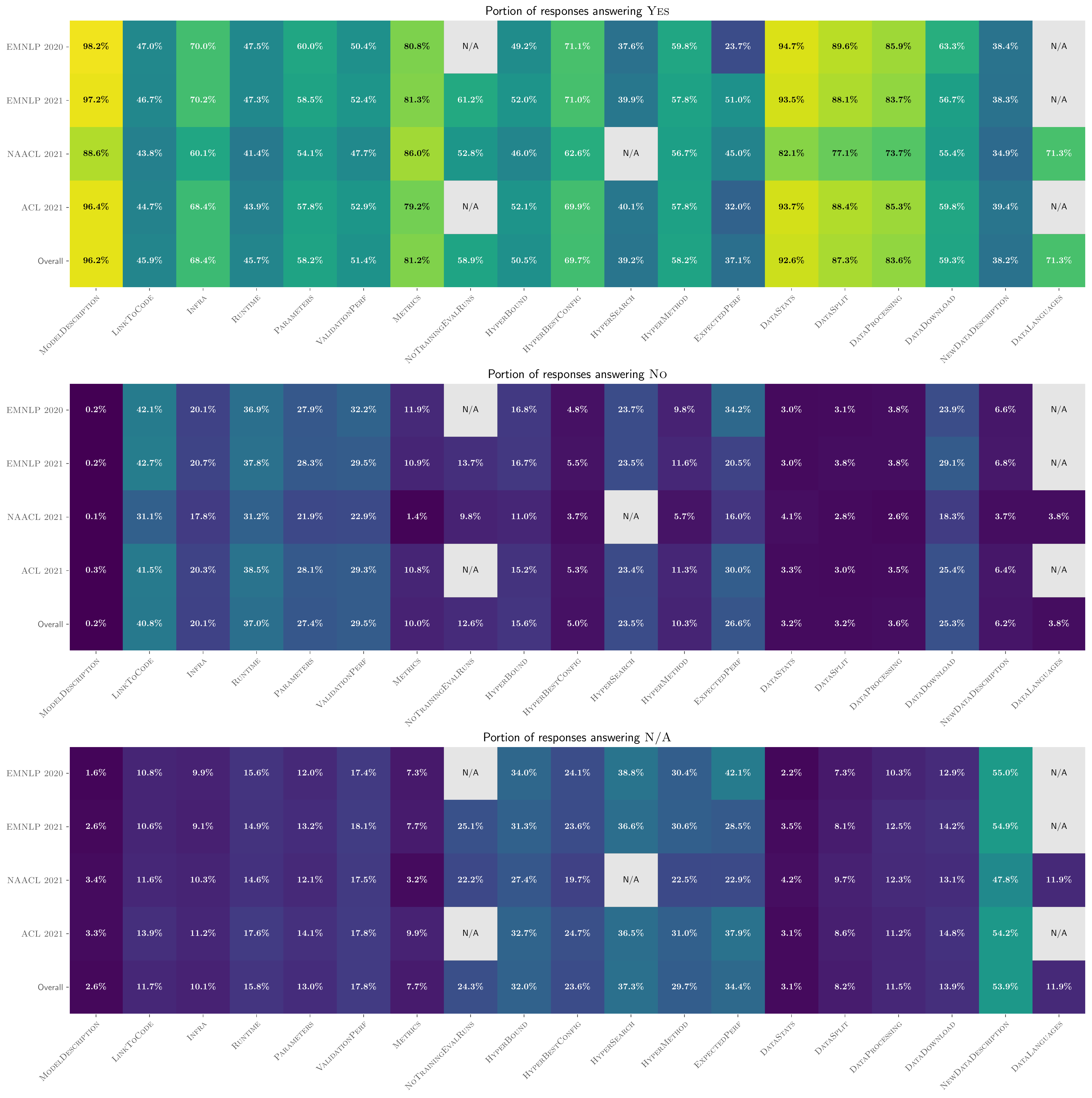}
\centering
\end{figure*}

\paragraph{\Yes\ was the most common response to checklist questions.}
The proportion of a given answer in responses for each question is shown in Figure~\ref{fig:portions_all_ans_per_q}. $\cent{62.65}$ of responses to checklist questions across all conferences were \Yes, with $\cent{62.77}$, $\cent{63.71}$, $\cent{59.97}$, and $\cent{62.46}$ respectively for \TheConfs. The majority of the questions have greater than $50\%$ \Yes\ response rate over all conferences. Only \LinkToCode, \Runtime, \HyperSearch, \Summary, and \NewData\ receive less than half \Yes\ responses. All questions receive more \Yes\ responses than \No\, and only \NewData\ receives more \NA\ than \Yes.

\begin{figure*}[t]
\caption{\Accept\ rates for submissions with a given response. Column (A) shows rate conditioned on response regardless of item. (B) conditions on answer and item. Rows present each conference and pooled results overall.}
\label{fig:accept_rates_by_q}
\includegraphics[width=\textwidth]{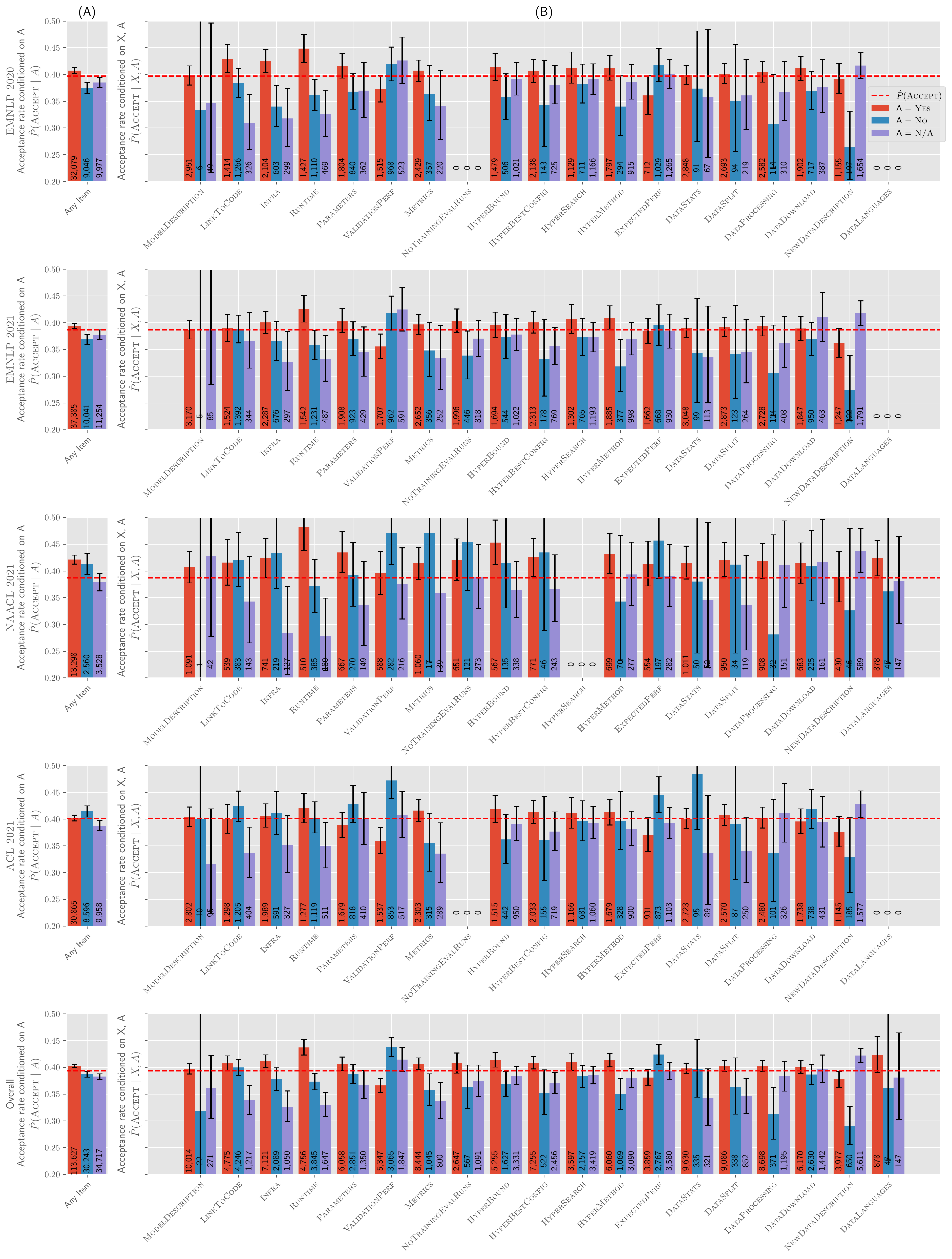}
\centering
\end{figure*}

% \jdcomment{Let's move the rest of this paragraph to the appendix, and say something like, ``Figure~\ref{fig:accept_rates_by_q_overall} shows the acceptance rate averaged across items, and Figure~\ref{something} in the appendix shows all checklist items. We have further discussion of these results in Appendix~\ref{something}.''}
\paragraph{The checklist items which receive less than average \Yes\ acceptance rates are not consistent across all conferences.} Figure~\ref{fig:accept_rates_by_q} shows acceptance rates for all checklist items over all conferences. From this figure we see that \acl\ also has \LinkToCode, \Parameters, \DataStats, and \DataDownload\ \Yes\ acceptance rates below average, though all of these estimates include the average acceptance rate within their 95\% confidence intervals. \naacl\ has no \Yes\ acceptance rates below average, though \ValidationPerf\ and \NewData\ remain the two lowest. Likely all \naacl\ \Yes\ acceptance rates are elevated because in this conference respondents could leave questions \Blank, possibly diverting some low-quality responses to \Blank\ instead of \Yes. Also of note, however is that across conferences \Runtime\ receives high \Yes\ acceptance rate, achieving the best overall at $\cent{4.349}$ higher than average.

% \paragraph{The ordering of acceptance rates by answers to a given question changes between conferences.} The overall results more closely reflect ordering for \emnlpTwenty\ and \emnlpTwentyOne, where only $3$ and $5$ questions have different orders, while \naacl\ and \acl\ have more different orderings, $10$ and $9$ respectively.
%For \emnlpTwenty\ and \emnlpTwentyOne\ There is a difference between \ExpectedPerf\  likely based on the difference in question wording where \emnlpTwenty\ had more stringent requirements. Three other questions saw changes in ordering but with substantial overlap in error bars. \naacl\ and \acl\ see higher acceptance rate for \No\ responses to\ValidationPerf\ while \emnlpTwenty\ and \emnlpTwentyOne\ saw the higher rate for \NA. 

\paragraph{There are outliers to the little variation in response proportions between conferences, but they are likely artifacts of changes in the checklist.} The rate of \Yes\ responses is generally lower for \naacl, but this is likely due to the ability to leave checklist responses \Blank. Excluding \naacl, the largest difference in \Yes\ rate ($\cent{27.29}$) occurs on \ExpectedPerf\ when this item changes phrasing substantially between \emnlpTwenty\ and \emnlpTwentyOne.

\begin{table}[htbp]
\centering 
\begin{tabular}{lrlr}
\toprule 

% \multicolumn{4}{c}{Lazy (27.89\%)}\\
% \midrule
% Runtime & \textcolor{red}{-0.43} &Parameters & \textcolor{red}{-0.37} \\
% Infra & \textcolor{red}{-0.34} &LinkToCode & \textcolor{red}{-0.33} \\
% ValidationPerf & \textcolor{red}{-0.30} \\ 
% \midrule
% \multicolumn{4}{c}{Open Source (10.77\%)}\\
% \midrule
% LinkToCode & \textcolor{green}{0.71} &DataDownload & \textcolor{green}{0.48} \\
% \midrule
% \multicolumn{4}{c}{Long running experiments (9.20\%)}\\
% \midrule
% ValidationPerf & \textcolor{red}{-0.57} &ExpectedPerf & \textcolor{red}{-0.48} \\
% Runtime & \textcolor{green}{0.47} &Infra & \textcolor{green}{0.40} \\
% \midrule
% \multicolumn{4}{c}{Validating without hyperparameters (8.06\%)}\\
% \midrule
% HyperMethod & \textcolor{red}{-0.50} &HyperBound & \textcolor{red}{-0.49} \\
% ValidationPerf & \textcolor{green}{0.48} &HyperBestConfig & \textcolor{red}{-0.32} \\

\multicolumn{4}{c}{Partially adopted practices (\cent{27.89})}\\
\midrule
Runtime & \textcolor{darkorange}{-\float{0.43}} &Parameters & \textcolor{darkorange}{-\float{0.37}} \\
Infra & \textcolor{darkorange}{-\float{0.34}} &LinkToCode & \textcolor{darkorange}{-\float{0.33}} \\
ValidationPerf & \textcolor{darkorange}{-\float{0.30}} &DataDownload & \textcolor{darkorange}{-\float{0.30}} \\
HyperBound & \textcolor{darkorange}{-\float{0.26}} &ExpectedPerf & \textcolor{darkorange}{-\float{0.26}} \\
HyperMethod & \textcolor{darkorange}{-\float{0.23}} \\
\midrule
\multicolumn{4}{c}{Open source (\cent{10.77})}\\
\midrule
LinkToCode & \textcolor{darkcyan}{\float{0.71}} &DataDownload & \textcolor{darkcyan}{\float{0.48}} \\
Parameters & \textcolor{darkorange}{-\float{0.28}} &ValidationPerf & \textcolor{darkorange}{-\float{0.20}} \\
\midrule
\multicolumn{4}{c}{Long running experiments (\cent{9.20})}\\
\midrule
ValidationPerf & \textcolor{darkorange}{-\float{0.57}} &ExpectedPerf & \textcolor{darkorange}{-\float{0.48}} \\
Runtime & \textcolor{darkcyan}{\float{0.47}} &Infra & \textcolor{darkcyan}{\float{0.40}} \\
\midrule
\multicolumn{4}{c}{Validating without hyperparameters (\cent{8.06})}\\
\midrule
HyperMethod & \textcolor{darkorange}{-\float{0.50}} &HyperBound & \textcolor{darkorange}{-\float{0.49}} \\
ValidationPerf & \textcolor{darkcyan}{\float{0.48}} &HyperBestConfig & \textcolor{darkorange}{-\float{0.32}} \\
Runtime & \textcolor{darkcyan}{\float{0.25}} &Parameters & \textcolor{darkcyan}{\float{0.24}} \\

\bottomrule

\end{tabular}
\caption{The top four components from running PCA on shared checklist items from four conferences, with percent variance explained in parentheses. Each component lists checklist items and their coefficients with magnitude $> \float{0.2}$.}
\label{tab:pca_table}

\end{table}
\paragraph{PCA analysis.} To identify clusters of checklist items that relate to each other, we take inspiration from similar analysis in \citet{NLPMetasurvey} and use principal component analysis (PCA) on all responses to shared checklist items across the four conferences. This results in 16 features, which we linearize as $\{\No\ \rightarrow\ -1, \NA\ / \Blank\ \rightarrow\ 0, \Yes\ \rightarrow\ 1\}$. We run PCA using \texttt{scikit-learn} version 1.1.1 \cite{scikit-learn} and find that the first 4 components cover $\cent{55.93}$ of the variance in the data. Table~\ref{tab:pca_table} shows these components and their coefficients with magnitude $> \float{0.2}$. The first component assigns weight all in one polarity to the checklist items with middling frequencies, highlighting practices where perhaps community norms have not settled. The second component shows the intuitive connection between \LinkToCode\ and \DataDownload. The third captures what might be particularly resource intensive experiments that emphasize efficiency metrics but prevent running many experiment repetitions. Lastly the fourth component puts hyperparameter items in opposition with \ValidationPerf, perhaps pointing towards work that adapts choices other than traditional hyperparameters to a validation set.

\begin{figure*}[t]
\caption{Phi coefficient of binary \Yes\ or not \Yes\ answer for each item to binary \Accept\ or not \Accept\ for each submission.}
\label{fig:phi_per_q_to_accept_findings}
\includegraphics[width=\textwidth]{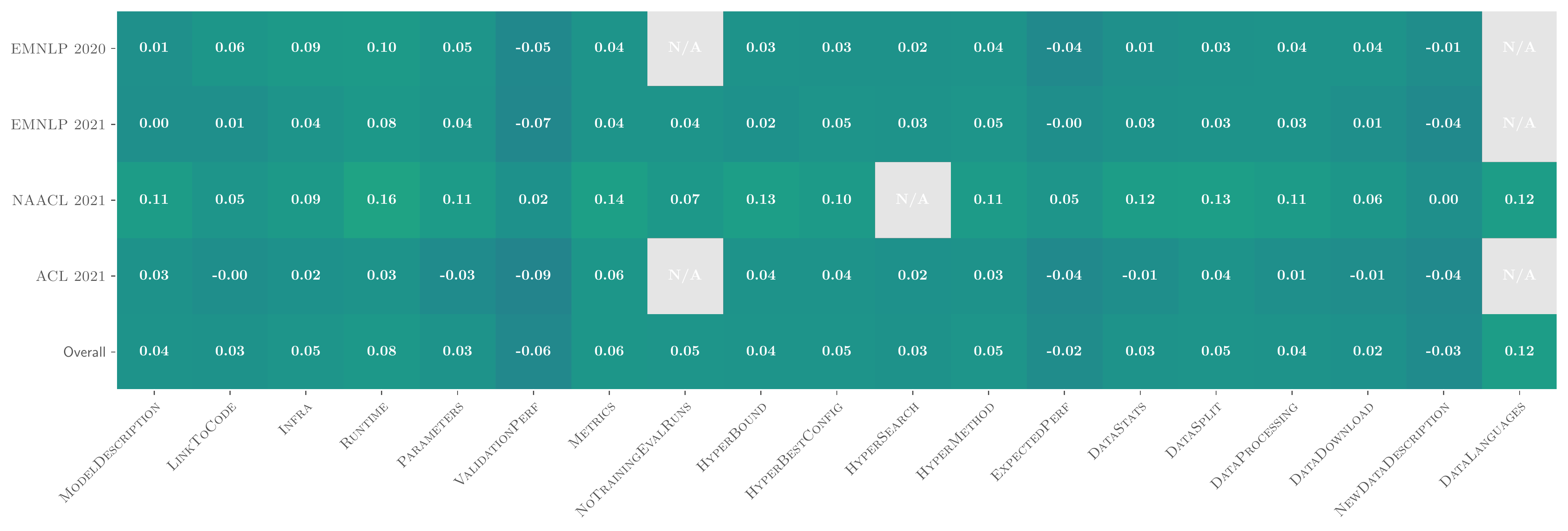}
\centering
\end{figure*}

% Figure \ref{fig:phi_per_q_to_accept_findings} shows the phi coefficient between whether or not a question was answered \Yes\ and acceptance to \Main\ or \Findings.

\begin{figure*}[t]
\caption{Phi coefficient between items shared over all conferences for the binary variable \Yes\ or not \Yes. Unsurprisingly, related groups of items about efficiency, hyperparameters, and data each correlate together.}
\label{fig:q_to_q_phi}
\includegraphics[width=\textwidth]{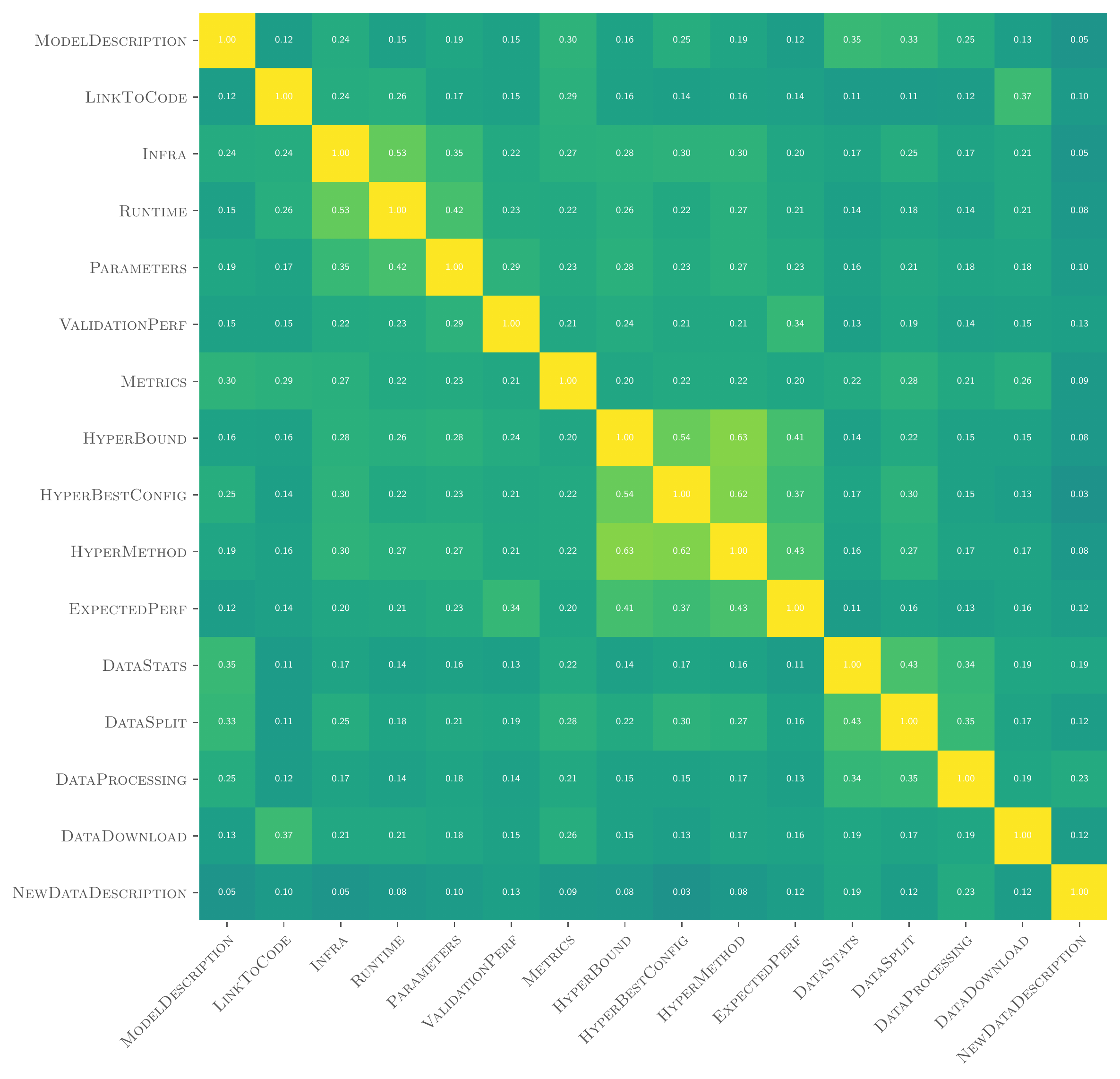}
\centering
\end{figure*}

% Figure \ref{fig:q_to_q_phi} \todo{Running text}

% Figure \ref{fig:portions_all_ans_per_q} \todo{Running text}

% \begin{figure*}[t]
% \caption{\Main\ + \Findings\ Acceptance rates for submissions with a given answer to a given question aggregated over all four conferences.}
% \label{fig:accept_rates_by_q_overall_FULL}
% \includegraphics[width=\textwidth]{figures/accept_rates_by_q_overall_FULL.png}
% \centering
% \end{figure*}

% Figure \ref{fig:accept_rates_by_q_overall_FULL} \todo{Running text}

% Figure \ref{fig:accept_rates_by_q} \todo{Running text}

% \begin{figure*}[t]
% \caption{Average reviewer assessed reproducibility score (from 1-5) for papers with a given answer to a given question aggregated over two conferences where this data was available.}
% \label{fig:reprod_per_answ_overall_FULL}
% \includegraphics[width=\textwidth]{figures/reprod_per_answ_overall_FULL.png}
% \centering
% \end{figure*}

% Figure \ref{fig:reprod_per_answ_overall_FULL} \todo{Running text}

\begin{figure*}[t]
\caption{Reviewer perceived reproducibility score (\Reprod $\in [1,5]$) for submissions with a given response. Column (A) shows score conditioned on response regardless of item. (B) conditions on answer and item. Rows present the two conferences with such data and pooled results overall.}
\label{fig:reprod_per_answ_per_q}
\includegraphics[width=\textwidth]{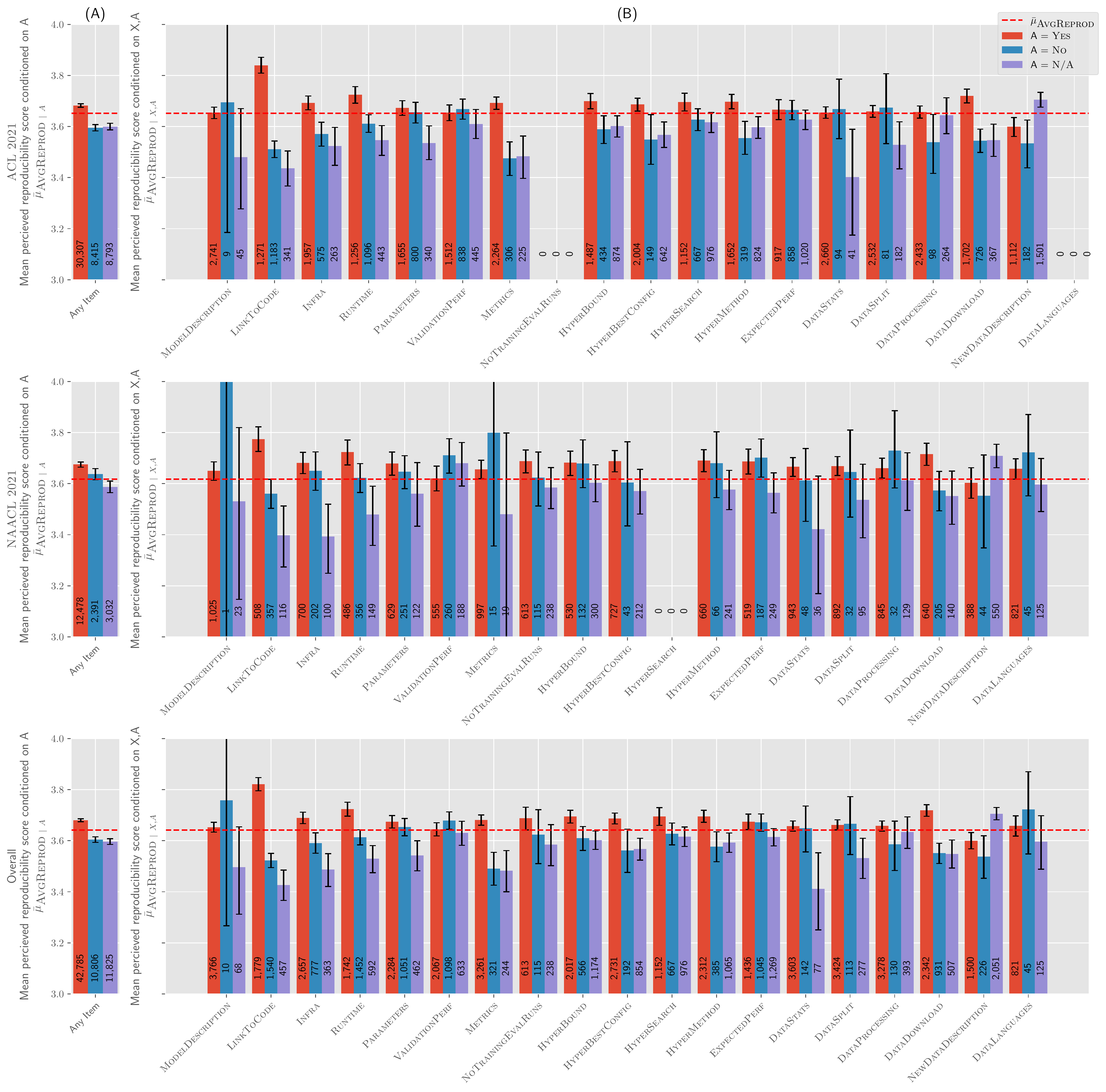}
\centering
\end{figure*}

% Figure \ref{fig:reprod_per_answ_per_q} \todo{Running text}

\begin{figure*}[t]
\caption{Proportion (row labels) and \Accept\ rates (horizontal purple bars) over all conferences for top response patterns for items split into three sections.
}
\label{fig:top_partial_resp_pattern_overall}
\includegraphics[width=\textwidth]{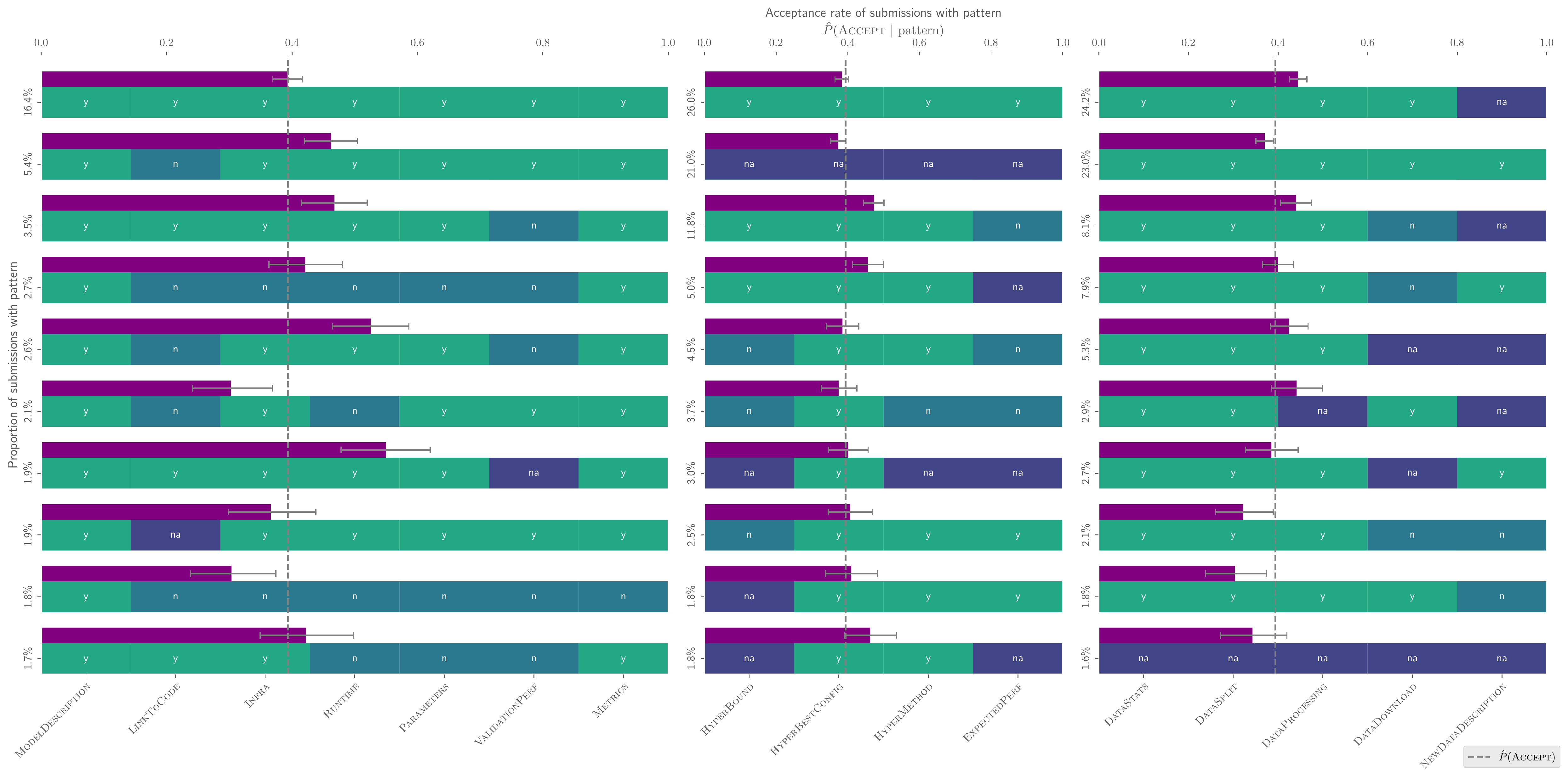}
\centering
\end{figure*}

% Figure \ref{fig:top_partial_resp_pattern_overall} \todo{Running text}
% Notes on top k
% Story beats
% Responding yes to everything puts you above avg accept rate (especially in first 2 sections)
% “Honest” response patterns have higher accept (cuz authors know their paper’s are good enough to be honest?) Such as no link to code
% Or no linktocode corresponds to larger, industry work?
% Non-reporting papers for “easy” items get dinged (such as runtime, newdata) 
% BUT runtime with link to code is higher!? -> runtime is important in absence of code because you can’t go look yourself.
% Also not having download link lessens impact of no data collection description
% N/a to validation performance is very high, perhaps because there was no tuning and so papers are more sciency and less engineeringy?

\begin{table*}[htbp]
\caption{Exact checklist item phrasing for each conference. Items listed as \NA\ did not appear on the checklist for that conference.}
\label{tab:exact_questions_table}
\tiny
\centering 
\begin{tabular}{lp{3cm}p{3cm}p{3cm}p{3cm}}
\toprule
        Short Name &                                                                                                                                                                              EMNLP 2020 &                                                                                                                                                                  EMNLP 2021 &                                                                                                                                                                       NAACL 2021 &                                                                                                                                                                    ACL 2021 \\ \midrule
  ModelDescription &                                                                                                               A clear description of the mathematical setting, algorithm, and/or model. &                                                                                                   A clear description of the mathematical setting, algorithm, and/or model. &                                                                                                         A clear description of the mathematical setting, algorithm, and/or model &                                                                                                    A clear description of the mathematical setting, algorithm, and/or model \\ \midrule
        LinkToCode &                                                                              A link to a downloadable source code, with specification of all dependencies, including external libraries & Submission of a zip file containing source code, with specification of all dependencies, including external libraries, or a link to such resources (while still anonymized) & A link to a downloadable source code, with specification of all dependencies, including external libraries (recommended for camera ready, though welcome for initial submission) & Submission of a zip file containing source code, with specification of all dependencies, including external libraries, or a link to such resources (while still anonymized) \\ \midrule
             Infra &                                                                                                                                            Description of computing infrastructure used &                                                                                                                                Description of computing infrastructure used &                                                                                                                                   A description of computing infrastructure used &                                                                                                                                Description of computing infrastructure used \\ \midrule
           Runtime &                                                                                                                                                       Average runtime for each approach &                                                                 The average runtime for each model or algorithm (e.g., training, inference, etc.), or estimated energy cost &                                                                                                        The average runtime for each model or algorithm, or estimated energy cost &                                                                                                                                           Average runtime for each approach \\ \midrule
        Parameters &                                                                                                                                                      Number of parameters in each model &                                                                                                                                          Number of parameters in each model &                                                                                                                                           The number of parameters in each model &                                                                                                                                          Number of parameters in each model \\ \midrule
    ValidationPerf &                                                                                                                      Corresponding validation performance for each reported test result &                                                                                                          Corresponding validation performance for each reported test result &                                                                                                               Corresponding validation performance for each reported test result &                                                                                                          Corresponding validation performance for each reported test result \\ \midrule
           Metrics &                                                                                                                              Explanation of evaluation metrics used, with links to code &                                                                                                                  Explanation of evaluation metrics used, with links to code &                                                                                      A clear definition of the specific evaluation measure or statistics used to report results. &                                                                                                                  Explanation of evaluation metrics used, with links to code \\ \midrule
NoTrainingEvalRuns &                                                                                                                                                                                     N/A &                                                                                                                            The exact number of training and evaluation runs &                                                                                                                                 The exact number of training and evaluation runs &                                                                                                                                                                         N/A \\ \midrule
        HyperBound &                                                                                                                                                          Bounds for each hyperparameter &                                                                                                                                              Bounds for each hyperparameter &                                                                                                                                               The bounds for each hyperparameter &                                                                                                                                              Bounds for each hyperparameter \\ \midrule
   HyperBestConfig &                                                                                                                                Hyperparameter configurations for best-performing models &                                                                                                                    Hyperparameter configurations for best-performing models &                                                                                                                     The hyperparameter configurations for best-performing models &                                                                                                                    Hyperparameter configurations for best-performing models \\ \midrule
       HyperSearch &                                                                                                                                                  Number of hyperparameter search trials &                                                                                                                                      Number of hyperparameter search trials &                                                                                                                                                                              N/A &                                                                                                                                      Number of hyperparameter search trials \\ \midrule
       HyperMethod &                                 The method of choosing hyperparameter values (e.g., uniform sampling, manual tuning, etc.) and the criterion used to select among them (e.g., accuracy) &                     The method of choosing hyperparameter values (e.g., uniform sampling, manual tuning, etc.) and the criterion used to select among them (e.g., accuracy) &                            The method of choosing hyperparameter values (e.g. manual tuning, uniform sampling, etc.) and the criterion used to select among them (e.g. accuracy) &                     The method of choosing hyperparameter values (e.g., uniform sampling, manual tuning, etc.) and the criterion used to select among them (e.g., accuracy) \\ \midrule
      ExpectedPerf & Expected validation performance, as introduced in Section 3.1 in * Dodge et al, 2019, or another measure of the mean and variance as a function of the number of hyperparameter trials. &                                                                                                  Summary statistics of the results (e.g., mean, variance, error bars, etc.) &                                                                                                        Summary statistics of the results (e.g. mean, variance, error bars, etc.) &                                                              Expected validation performance, or the mean and variance as a function of the number of hyperparameter trials \\ \midrule
         DataStats &                                                                                                                                          Relevant statistics such as number of examples &                                                                                          Relevant details such as languages, and number of examples and label distributions &                                                                                                           Relevant statistics such as number of examples and label distributions &                                                                                                                              Relevant statistics such as number of examples \\ \midrule
         DataSplit &                                                                                                                                                 Details of train/validation/test splits &                                                                                                                                     Details of train/validation/test splits &                                                                                                                                          Details of train/validation/test splits &                                                                                                                                     Details of train/validation/test splits \\ \midrule
    DataProcessing &                                                                                                                Explanation of any data that were excluded, and all pre-processing steps &                                                                                                    Explanation of any data that were excluded, and all pre-processing steps &                                                                                                      An explanation of any data that were excluded, and all pre-processing steps &                                                                                                    Explanation of any data that were excluded, and all pre-processing steps \\ \midrule
      DataDownload &                                                                                                                                            A link to a downloadable version of the data &                                                                                                    A zip file containing data or link to a downloadable version of the data &                                                                                                        A link to a downloadable version of the dataset or simulation environment &                                                                                                                                A link to a downloadable version of the data \\ \midrule
           NewDataDescription &                                      For new data collected, a complete description of the data collection process, such as instructions to annotators and methods for quality control. &                          For new data collected, a complete description of the data collection process, such as instructions to annotators and methods for quality control. &                                For new data collected, a complete description of the data collection process, such as instructions to annotators and methods for quality control &                           For new data collected, a complete description of the data collection process, such as instructions to annotators and methods for quality control \\ \midrule
     DataLanguages &                                                                                                                                                                                     N/A &                                                                                                                                                                         N/A &                                                                                                                           For natural language data, the name of the language(s) &                                                                                                                                                                         N/A \\
\bottomrule
\end{tabular}
\end{table*}

\subsection{Baseline Acceptance Rates}
\label{app:baseline_acceptance_rates}
\begin{table*}[!htbp]
\caption{Submissions and decisions statistics. Reported acceptance rates include varying amounts of withdrawn and desk-reject submissions. We exclude all of these to standardize to rates.}
\label{tab:reported_vs_our_data}
\centering 
\begin{tabular}{llrrrr}
\toprule
 &   &  & \multirow[b]{2}{*}{\parbox{20mm}{Withdrawn /\\ Desk-Reject}} & \multicolumn{2}{c}{Accept Rate}\\
 \cmidrule{5-6}
& Conference & Submissions & & \Main & \Findings\\
\midrule 
\multirow{4}{*}{\rotatebox[]{90}{Reported}}
& EMNLP 2020 &
  3359 &
  - &
  \cent{22.39} &
  -\\
& EMNLP 2021 &
  3600 &
  - &
  \cent{23.33} &
  \cent{11.64}\\
& NAACL 2021 &
  1797 &
  - &
  \cent{26.54} &
  N/A\\
  & ACL 2021 &
  3350 &
  - &
  \cent{21.19} &
  \cent{13.64} \\
\midrule 
% \multirow{4}{*}{\rotatebox[origin=lc]{90}{\parbox{16mm}{Our Data \\ \footnotesize{(Excluding \\ Withdrawn)}}}}
\multirow{5}{*}{\rotatebox[origin=c]{90}{Our Data}}
& EMNLP 2020 &
  3666 &
  660 &
  \cent{24.92} &
  \cent{14.80}\\
& EMNLP 2021 &
  4815 &
  1555 &
  \cent{25.80} &
  \cent{12.85} \\
& NAACL 2021 &
  1797 &
  565 &
  \cent{38.72} &
  N/A\\
& ACL 2021 &
  3377 &
  470 &
  \cent{24.42} &
  \cent{15.72}\\
& Overall &
  13655 &
  3250 &
  \multicolumn{2}{c}{\cent{39.38}}\\

% \midrule 
% Also with all same answers removed
% \parbox[t]{2mm}{\multirow{3}{*}{\rotatebox[origin=c]{90}{Cleaned}}}
% & EMNLP 2020 &
%   2854 &
%   25.23\% &
%   15.00\%\\
% & EMNLP 2021 &
%   3000 &
%   26.37\% &
%   12.80\%\\
% & NAACL 2021 &
%   1049 &
%   40.80\% &
%   N/A\\
% & Overall &
%   6903 &
%   2751 (39.85\%) &
%   1939 (28.09\%) &
%   812 (11.76\%)\\
\bottomrule 
\end{tabular} 
\end{table*}
To aid in analyzing how publication decisions differ based on responses to the checklist, we first must establish what is the average acceptance rate across all papers in our data. 
In Table~\ref{tab:reported_vs_our_data} we provide basic statistics about submissions and decisions. 
The acceptance rates reported\footnote{\url{https://aclweb.org/aclwiki/Conference\_acceptance\_rates}} by the conferences all include an unknown and varying number of withdrawn and desk-rejected papers and thus are not easily comparable. 
For the rest of our analysis we will instead make use of acceptance rates computed from our data that always remove all withdrawn and desk-rejected papers. 
With this approach we find that all of the conferences have similar acceptance rates when including both acceptance to the \Main\ conference and to \Findings.

\end{document}